\documentclass{sig-alternate}
\pdfoutput=1

\usepackage{amsmath,amssymb}

\usepackage{array,multirow}

\usepackage{graphicx} 
\usepackage{subfigure} 

\usepackage{natbib}
\usepackage{algorithm}
\usepackage[noend]{algorithmic-kr}

\usepackage[hyphens]{url}

\usepackage{enumitem}
\usepackage{mdwlist}

\newcommand{\argmax}{\texttt{argmax} }
\newcommand{\nn}{\nonumber }
\newcommand{\nnp}{\nonumber \\ }
\newcommand{\ie}{{\it i.e., }}
\newcommand{\eg}{{\it e.g., }}
\newcommand{\expct}{\mathbf E}

\newcommand{\assignments}{D}
\newcommand{\assignment}{d}
\newcommand{\graders}{G}
\newcommand{\grader}{g}
\newcommand{\graderIngraders}{\grader \in \graders}
\newcommand{\graderassign}[1]{\assignments_{#1}}
\newcommand{\graderassignspec}{\graderassign{\grader}}
\newcommand{\sample}{S}
\newcommand{\cardinalgrade}[2]{y^{(#1)}_{#2}}
\newcommand{\orderinggrade}[1]{\sigma^{(#1)}}
\newcommand{\orderinggradespec}{\orderinggrade{\!\grader\!}\!}
\newcommand{\orderingtrue}{\sigma^*}
\newcommand{\ordering}{\sigma}
\newcommand{\orderingmean}{\bar{\sigma}}
\newcommand{\orderingest}{\hat{\sigma}}
\newcommand{\cardinalvec}{s}
\newcommand{\cardinalestvec}{\hat{s}}
\newcommand{\cardinal}[1]{s_{#1}}
\newcommand{\cardinaltrue}[1]{s^*_{#1}}
\newcommand{\cardinalest}[1]{\hat{s}_{#1}}
\newcommand{\graderreliability}[1]{\eta_{#1}}
\newcommand{\graderreliabilityest}[1]{\hat{\eta}_{#1}}
\newcommand{\graderreliabilityvec}{\eta}
\newcommand{\graderreliabilityvecest}{\hat{\eta}}
\newcommand{\graderreliabilityspec}{\graderreliability{\grader}}
\newcommand{\graderreliabilityspecest}{\graderreliabilityest{\grader}}
\newcommand{\tauerror}{\mathcal{E}_K}
\newcommand{\deltakerror}{\delta_K}
\newcommand{\deltaskerror}{\delta_{SK}}
\newcommand{\deltakterror}{\tau_{KT}}
\newcommand{\score}[1]{s_{#1}}

\newcommand{\preferencesgrad}[1]{\rho^{(#1)}}
\newcommand{\preferencesgradspec}{\preferencesgrad{\grader}}
\newcommand{\preferredto}[1]{\succ_{\!#1}}

\newcommand{\BlackBox}{\rule{1.5ex}{1.5ex}}  

\newtheorem{defn}{Definition}

{\makeatletter
 \gdef\xxxmark{%
   \expandafter\ifx\csname @mpargs\endcsname\relax 
     \expandafter\ifx\csname @captype\endcsname\relax 
       \marginpar{xxx}
     \else
       xxx 
     \fi
   \else
     xxx 
   \fi}
 \gdef\xxx{\@ifnextchar[\xxx@lab\xxx@nolab}
 \long\gdef\xxx@lab[#1]#2{{\bf [\xxxmark #2 ---{\sc #1}]}}
 \long\gdef\xxx@nolab#1{{\bf [\xxxmark #1]}}
}

\usepackage{xcolor,colortbl}

\makeatletter
\newcommand{\thickhline}{%
    \noalign {\ifnum 0=`}\fi \hrule height 1pt
    \futurelet \reserved@a \@xhline
}
\makeatother

\newcommand{\justforref}[1]{}
\newcommand{\longversion}[1]{}

\newcommand{\shortversion}[1]{#1}

\usepackage{collcell}
\makeatletter
\newcolumntype{G}{>{\collectcell\@gobble}c<{\endcollectcell}@{}}
\newcolumntype{E}[1]{>{\collectcell\@gobble}p{0in}<{\endcollectcell}@{}}
\makeatother

\setlist[enumerate]{topsep=2pt,itemsep=-0.5ex,partopsep=1ex,parsep=1ex}
\setlist[itemize]{topsep=2pt,itemsep=-0.5ex,partopsep=1ex,parsep=1ex}

 \setdescription{leftmargin=0.5cm,labelindent=0cm}

\begin{document} 
%
\conferenceinfo{KDD}{'14 New York City, NY USA}

\title{Methods for Ordinal Peer Grading}

\numberofauthors{2} 

\author{
\alignauthor
Karthik Raman\\
       \affaddr{Department of Computer Science}\\
       \affaddr{Cornell University, Ithaca NY 14853}\\
       \email{karthik@cs.cornell.edu}
\alignauthor
Thorsten Joachims\\
       \affaddr{Department of Computer Science}\\
       \affaddr{Cornell University, Ithaca NY 14853}\\
       \email{tj@cs.cornell.edu}
}

\maketitle

\begin{abstract}
MOOCs have the potential to revolutionize higher education with their wide outreach and accessibility, but they require instructors to come up with scalable alternates to traditional {\em student evaluation}.
Peer grading --  having students assess each other  -- is a promising approach to tackling the problem of evaluation at scale, since the number of "graders" naturally scales with the number of students. 
However, students are not trained in grading, which means that one cannot expect the same level of grading skills as in traditional settings. 
Drawing on broad evidence that {\em ordinal} feedback is easier to provide and more reliable than {\em cardinal} feedback \cite{ barnett_2003,stewart_2005,miller_1956, carterette_2008}, it is therefore desirable to allow peer graders to make ordinal statements (e.g. "project X is better than project Y") and not require them to make cardinal statements (e.g. "project X is a B-").
Thus, in this paper we study the problem of automatically inferring student grades from ordinal peer feedback, as opposed to existing methods that require cardinal peer feedback.
We formulate the ordinal peer grading problem as a type of rank aggregation problem, and explore several probabilistic models under which to estimate student grades and grader reliability.
We study the applicability of these methods using peer grading data collected from a real class --- with instructor and TA grades as a baseline --- and demonstrate the efficacy of ordinal feedback techniques in comparison to existing cardinal peer grading methods.
Finally, we compare these peer-grading techniques to traditional evaluation techniques.
\end{abstract}

\category{H.4}{Information Systems Applications}{Miscellaneous}

\terms{Algorithms, Experimentation, Theory}

\keywords{Peer Grading, Ordinal Feedback, Rank Aggregation}

\section{Introduction} 
\label{sec:intro}

The advent of {\em MOOCs} (Massive Online Open Courses) promises unprecedented access to education given their relatively low costs and broad reach, empowering learning across a diverse range of subjects for anyone with access to the Internet.
Classes frequently have upwards of 20000 students,
which is orders of magnitude larger than a conventional university class.
Thus instructors are forced to rethink classroom logistics and practices so as to scale to MOOCs.

One of the key open challenges is {\em student evaluation} for such large classes.
Traditional assessment practices, such as instructors or teaching assistants (TAs) grading individual student assignments, are simply infeasible at this scale.
Consequently assignments in most current MOOCs take the form of simple multiple-choice questions and other schemes that can be graded automatically.
However, relying on such rigid testing schemes does not necessarily serve as a good indicator of learning and falls short of conventional test-design standards \cite{jonathanhaber_assessment,jonathanhaber_screwing}. Furthermore, such a restrictive testing methodology limits the learning outcomes that can be tested, or even limit the kinds of courses that can be offered. For example, liberal-arts courses and research-oriented classes require more open-ended assignments and responses (e.g., essays, project proposals, project reports).

{\em Peer grading} has the potential to overcome the limitations outlined above while scaling to the size of even the largest MOOCs.
In peer grading, students --- not instructors or TAs --- provide feedback on the work of other students in their class \cite{freeman_2010,kulkarni:2013},
meaning that the number of ``graders'' naturally grows with the number of students.
While the scaling properties of peer grading are attractive, there are several challenges in making peer grading work. 


One key challenge lies in the fact that students are not trained graders, which argues for making the feedback process as simple as possible. Given broad evidence that for many tasks {\em ordinal} feedback is easier to provide and more reliable than {\em cardinal} feedback \cite{barnett_2003,stewart_2005,miller_1956, carterette_2008}, it is therefore desirable to base peer grading on ordinal feedback (e.g. "project X is better than project Y"). Unfortunately, all existing methods for aggregating peer feedback into an overall assessment require that students provide {\em cardinal} feedback (e.g. "project X is a B-"). Furthermore, the efficacy of simple techniques for aggregating cardinal feedback, such as averaging, has been questioned \cite{bouzidi_2009,chang_2011,mosfert_2013}. While probabilistic machine learning methods have recently been proposed to address these challenges \cite{piech_peergrading_2013}, they still face the problem that students may be grading on different scales. For example, students may have a preconception of what constitutes a B+ based on the university they come from. These scales may also be non-linear as the difference between an A+ and an A may not be the same as the difference between a C+ and a C.

To overcome the problems of cardinal feedback, we introduce the task of {\em ordinal peer grading} in this paper. By having students give ordinal statements and not cardinal statements as feedback, we offload the problem of developing a scale from the student onto the peer grading algorithm. 
The key technical contributions of this paper lie in the development of methods for ordinal peer grading, where the goal is to automatically infer an overall assessment of a set of assignments from ordinal peer feedback. Furthermore, a secondary goal of our methods is to infer how accurately each student provides feedback, so that reliable grading can be incentivized (e.g., as a component of the overall grade). 
To this effect, we propose several machine learning methods for {\em ordinal peer grading}, which differ by how probability distributions over rankings are modeled. 
For these models, we provide efficient algorithms for estimating assignment grades and grader reliabilities.

To study the applicability of our methods in real-world settings, we collected peer assessment data as part of a university-level course.
Using this data, we demonstrate the efficacy of the proposed ordinal feedback techniques in comparison to the existing cardinal feedback techniques.
Furthermore, we compare our ordinal peer grading methods with traditional evaluation techniques that were used in the course in parallel.
Using this classroom data we also investigate other properties of these techniques, such as their robustness, data dependence and self-consistency.
Finally, we analyze the responses to a survey completed by students in the classroom experiment, indicating that most students found the peer grading experience (receiving and providing feedback) helpful and valuable.

\section{The Peer Grading Problem}
\label{sec:problem}

We begin by formally defining the peer grading problem, as it presents itself from a machine learning perspective. We are given a set of $|\assignments|$ {\em assignments} $\assignments=\{\assignment_1,...,\assignment_{|\assignments|}\}$ (\eg essays, reports) which need to be graded.
Grading is done by a set of $|\graders|$ graders $\graders=\{\grader_1,...,\grader_{|\graders|}\}$ (\eg student peer grader, reviewers), where each grader receives a subset $\graderassign{\grader} \subset \assignments$ to assess. The choice of assignments for each grader can be uniformly random, or can follow a deterministic or sequential design. In either case, the number of assignments that any grader assesses $|\graderassign{\grader}|$ is much smaller than the total number of assignments $|\assignments|$ (\eg |$\graderassign{\grader}| \approx 10$).

Each grader provides feedback for his or her set of assignments $\graderassign{\grader}$. Ordinal and cardinal peer grading differ in the type of feedback a grader is expected to give:
\begin{description}
 \item[Cardinal Peer Grading (CPG):] In cardinal peer grading, each grader $\grader$ provides cardinal-valued feedback for each item $\assignment \in \graderassign{\grader}$. Typically, this is a numeric or categorical response which we denote as $\cardinalgrade{\grader}{\assignment}$ (\eg Likert scale, letter grade).
 \item[Ordinal Peer Grading (OPG):] In ordinal peer grading, each grader $\grader$ returns an ordering $\orderinggrade{\grader}$ (possibly with ties) of his or her assignments $\graderassign{\grader}$, indicating relative but not absolute quality. More generally, ordinal feedback could also consist of multiple pairwise preferences, but we focus on the case of a single ordering in this paper.
\end{description}
Independent of the type of feedback that graders provide, the goal in peer grading is twofold. 

We call the first goal {\em grade estimation}, which is the task of estimating the true quality of the assignments in $\assignments$ from the grader feedback. We distinguish between two types of grade estimation, which differ by how they express assignment quality. In {\em ordinal grade estimation}, the goal is to infer a ranking $\hat{\sigma}$ of all assignments in $\assignments$ that most accurately reflects some true ordering (by quality) $\orderingtrue$. In {\em cardinal grade estimation}, the goal is to infer a cardinal grade $\cardinalest{\assignment}$ for each $\assignment \in \assignments$ that most accurately reflects each true grade $\cardinaltrue{\assignment}$. Note that the type of feedback does not necessarily determine whether the output of grade estimation is ordinal or cardinal. In particular, we will see that some of our methods can infer cardinal grades even if only given ordinal feedback.

The second goal is {\em grader reliability estimation}, which is the task of estimating how accurate the feedback of a grader is. Estimating grader reliability is important for at least two reasons. First, identifying unreliable grades allows us to downweight their feedback for grade estimation. Second, and more importantly, it allows us to incentivize good and thorough grading by making peer grading itself part of the overall grade. In the following, we will typically represent the reliability of a grader as a single number $\graderreliability{\grader} \in \Re^+$.

In the following sections, we derive and evaluate methods for grade estimation and grader reliability estimation in the Ordinal Peer Grading setting.

\begin{table}[t]
\centering
\footnotesize
 \begin{tabular}{|c| p{2.48in}|}
 \hline
 \longversion{
 $G,g (\in G)$ & Set of all graders, Specific grader \\
 $D,d (\in D)$ & Set of all items, Specific item \\
 }
 $D_g (\subset D)$ & Set of items graded by grader $g$ \\
 $s_d (\in \Re)$ & Predicted grade for item $d$ (larger is better) \\
 $\eta_g (\in \Re^+)$ & Predicted reliability of grader $g$ \\
 $\sigma_g$ & Ranking feedback (with possible ties) from $g$ \\
 $r_d^{(\sigma)}$ & Rank of item $d$ in ranking $\sigma$ (rank 1 is best) \\
 $\rho_g$ & Set of pairwise preference feedback from $g$ \\
 $\!d_2\!\succ_{\sigma}\!d_1\!\!$ & $d_2$ is preferred/ranked higher than $d_1$ (in $\sigma$) \\
 $\pi(A)$ & Set of all rankings over $A \subseteq D$ \\
$\sigma_1 \sim \sigma_2$ & $\exists$ way of resolving ties in $\sigma_2$ to obtain $\sigma_1$\\
 \hline
\end{tabular}
\vspace{-0.05in}
\caption{\label{tab:notation_used} Notation overview and reference.}
\vspace{-0.1in}
\end{table}

\subsection{Related Work in Rank Aggregation}
\label{sec:relatedwork}

The grade estimation problem in Ordinal Peer Grading can be viewed as a specific type of rank aggregation problem. Rank aggregation describes a class of problem related to combining the information contained in rankings from multiple sources.
Many popular methods used today \cite{guiver_placlucebayes_2009,lu_mallows_2011,chen_rankaggreg_2013} build on classical models and techniques such as the seminal work by Thurstone \cite{thurstone_1927}, Mallows \cite{mallows_1957}, Bradley \& Terry \cite{bradley_1952},
Luce \cite{luce_1961} and Plackett \cite{plackett_1975}.
These techniques have been used in different domains, each of which have branched off their own set of methods.

{\bf Search Result Aggregation} (also known as {\bf Rank Fusion} or {\bf Metasearch})
has the goal of merging search result rankings from different sources to produce a single output ranking. 
Such aggregation has been widely used to improve over the performance of any single ranker in both supervised and unsupervised settings \cite{aslam_metasearch_2001,qin_cpsrankagg_2010,volkovs_prefaggr_2012,niu_stochrankagg_2013}. 
Rank aggregation for search differs from Ordinal Peer Grading in several aspects. First, grader reliability estimation is not a goal in itself. Second, the success of search result aggregation depends mostly on correctly identifying the top items, while grade estimation aims to accurately estimate the full ranking. Third, ties and data sparsity are not an issue in search result aggregation, since (at least in principle) input rankings are total orders over all results.

{\bf Social Choice} and {\bf Voting Systems} perform rank aggregation on preferences that a set of individuals stated over competing items/interests/candidates. The goal is to identify the most preferred alternatives given conflicting preferences \cite{arrow_socialchoice_1970}. Commonly used  aggregation techniques are the {\em Borda count} and other Condorcet voting schemes \cite{aslam_metasearch_2001,dwork_rankagg_2001,lu_socialchoice_2010}.
These methods are ill-suited for the OPG problem, as they do not model voter reliability, typically assume rankings of all alternatives (or at least leave the choice of alternatives up to the voter), and usually focus on the top of the rankings.

{\bf Crowdsourcing} is probably the most closely related application domain, where the goal is to merge the feedback from multiple {\em crowdworkers} \cite{ipeirotis_crowdsourcing_2011,bashir_rankaggreg_2013}.
Due to the differing quality of these workers, modeling the worker reliability is essential \cite{raykar_learningfromcrowds_2010,chen_rankaggreg_2013}.
The key difference in our setting is that the number of items is large and we would like to correctly order all of them, not just identify the top-few.

Rank-aggregation has also been used for other settings such as multilabel/multiclass classification (by combining different classifiers) \cite{lebanon_cranking_2002} or for learning player skills in a gaming environment \cite{herbrich_trueskill_2007}.
Is is impossible to survey the vast literature on this topic and thus we refer the interested reader to a comprehensive survey on the topic \cite{liu_learningtorank_2009}.
These techniques have also been adapted for educational assessment \cite{bachrach_testgrade_2012}, via a graphical model based approach, for modeling the difficulty of questions and estimating the correct answers in a crowdsourced setting.
However these techniques are neither applicable for a peer grading setting nor can they handle open-ended answers (like essays).

\setlength{\textfloatsep}{10pt}
\begin{algorithm}[t]
\caption{{\bf Normal Cardinal-Score (NCS)} Algorithm (called {\bf PG}$_1$ in \cite{piech_peergrading_2013}) is used as a baseline in our experiments}
\label{alg:ncs_baseline}
\begin{algorithmic}
\STATE $s_d \sim \mathcal{N}(\mu_0,\frac{1}{\gamma_0})$ \hfill \COMMENT{True Scores}
\STATE $\eta_g \sim Gamma(\alpha_0,\beta_0)$ \hfill \COMMENT{Grader Reliability}
\STATE $b_g \sim \mathcal{N}(0,\frac{1}{\gamma_1})$ \hfill \COMMENT{Grader Bias (Only for NCS+G)}
\STATE $y_d^{(g)} \sim \mathcal{N}(s_d + b_g,\frac{1}{\eta_g})$ \hfill \COMMENT{Observed Cardinal Peer Grade}
\STATE Estimate $\cardinalest{\assignment},\graderreliabilityspecest$ and $\hat{b}_g$ \hfill \COMMENT{Using MLE}
\end{algorithmic}
\end{algorithm}

\subsection{Related Work in Peer Grading}

With the advent of online courses, peer grading has been increasingly used for large classes with mixed results \cite{bouzidi_2009,chang_2011,mosfert_2013}.
While most previous uses of peer grading have relied on simple estimation techniques like averaging cardinal feedback scores, recently a probabilistic learning algorithms has been proposed for peer grade estimation \cite{piech_peergrading_2013}. However, this method requires that students provide {\em cardinal} scores as grades.
A second limitation of the method in \cite{piech_peergrading_2013} is that they incentivize grader reliability by relating it to the grader's own assignment score.
However, such a setup is inappropriate when there are groups (such as our setting) or where external graders/reviewers are used (\eg conference reviewing). In addition, such an indirect incentive is harder to communicate and justify compared to the direct grader reliability estimates used in our case.
Lastly their approach requires that each student grades some assignments that were previously graded by the instructor in order to estimate grader reliability. This seems wasteful, given that students are only able to grade a small number of assignments in total.
We empirically compare their cardinal peer grading technique (Algorithm~\ref{alg:ncs_baseline}, using MLE instead of Gibbs sampling) with the ordinal peer grading techniques proposed in this paper.

Overall, given the limited amount of attention that the peer grading problem has received in the machine learning literature so far, we believe there is ample opportunity to improve on the state-of-the-art and address shortcomings that currently exist \cite{jonathanrees_peergradingcriticism}, which is reinforced by concurrent work on the topic by others \cite{diaz_nipsworkshop_2013,shah_2013}.


\section{Ordinal Peer Grading Methods}
\label{sec:methods}

In this section, we develop ordinal peer grading methods for grade estimation and then extend these methods to the problem of grader reliability estimation. Our methods are publicly available as software at \url{www.peergrading.org}, where we also provide a web service for peer grade estimation. 
These methods require as data an i.i.d. sample of orderings
\begin{eqnarray}
\sample=(\orderinggrade{\grader_1}, ..., \orderinggrade{\grader_{|G|}}),
\end{eqnarray}
where each ordering sorts a subset of assignment according to the judgment of grader $\grader_i$.


\subsection{Grade Estimation}\label{sec:grade_estimation}

Our grade estimation methods are based on models that represent probability distributions over rankings. In particular, we extend Mallow's Model (Sec~\ref{sec:mallowsmodel}), the Bradley-Terry model (Sec~\ref{sec:logisticpairmodel}), Thurstone's model (Sec~\ref{sec:thurstonemodel}), and the Plackett-Luce model (Sec~\ref{sec:plackettlucemodel}) as appropriate for the ordinal peer grading problem.

\subsubsection{Mallows Model (MAL and MALBC)}
\label{sec:mallowsmodel}

Mallow's model \cite{mallows_1957} describes a distribution over rankings $\ordering$ in terms of the distance $\delta(\orderingmean,\ordering)$ from a central ranking $\orderingmean$, which in our setting is the true ranking $\orderingtrue$ of assignments by quality.
\begin{equation}
P(\ordering|\orderingmean) = \frac{ e^{- \delta(\orderingmean,\ordering)}}{\sum_{\ordering'} e^{- \delta(\orderingmean,\ordering')}} \label{eq:mallowsmodel_probability}  
\end{equation}
While maximum likelihood estimation of $\orderingtrue$ given observed rankings is NP-hard for many distance functions \cite{dwork_rankagg_2001,qin_cpsrankagg_2010}, tractable approximations are known for special cases.
In this work, we use the following tractable {\bf Kendall-$\tau$ distance} \cite{kendall_rankcorrelation_1948}, which assumes that both rankings are total orderings over all assignments.
\begin{defn}
 We define the Kendall-$\tau$ Distance $\deltakerror$ between ranking $\ordering_1$ and ranking $\ordering_2$ as
 \begin{equation}
\deltakerror(\ordering_1,\ordering_2) = \sum_{\assignment_1 \preferredto{\ordering_1} \assignment_2} \! \mathbb{I}[[\assignment_2 \preferredto{\ordering_2} \assignment_1]] \label{eq:kendalltau_delta}
\end{equation}
\end{defn}
It measures the number of incorrectly ordered pairs between the two rankings. In our case, the rankings that students provide can have ties. We interpret these ties as {\em indifference} (\ie agnostic to either ranking), which leads to the following model, where the summation in the numerator is over all total orderings $\ordering'$ consistent with the weak ordering $\ordering$.
\begin{equation}
P(\ordering|\orderingmean) = \frac{ \sum_{\ordering' \sim \ordering}e^{- \delta(\orderingmean,\ordering')}}{\sum_{\ordering'} e^{- \delta(\orderingmean,\ordering')}} \label{eq:mallowsmodel_probability_ties}  
\end{equation}
Note also that the input ranking $\ordering$ may only sort a subset of assignments. In such cases, we appropriately restrict the normalization constant in (\ref{eq:mallowsmodel_probability_ties}). For Kendall-$\tau$ distance, this normalization constant can be computed efficiently, and it only depends on the number of elements in the ranking.
\begin{align}
\!\!Z_{M}(k)\!\!=\!\!\prod_{i=1}^{k}\!\!\Big(1\!+\!e^{-1}\!+\!\cdots\!+\!e^{-(i-1)}\!\Big)\!\!=\!\! \prod_{i=1}^{k} \frac{1 - e^{-i}}{1 - e^{-1} }\! \nn
\!
\end{align}
The numerator can likewise be computed efficiently.
Note that ties in the grader rankings $\orderinggradespec$ do not affect the normalization constant under the interpretation of indifference.

Under this modified Mallow's model, the maximum likelihood estimator of the central ranking $\orderingest$ is
\begin{equation}
\orderingest = \argmax_\ordering \left\{ \prod_{\graderIngraders}\!\frac{ \sum_{\ordering' \sim \orderinggradespec}e^{- \deltakerror(\ordering,\ordering')}  }{Z_{M}(|\graderassignspec|)} \right\}. \label{eq:mallowsmodel_likelihood}  
\end{equation}
Computing the maximum likelihood estimate $\orderingest$ as an estimate of the true ranking by quality $\orderingtrue$ requires finding the {\em Kemeny-optimal aggregate},
which is known to be NP-hard \cite{dwork_rankagg_2001}. However numerous approximations have been studied in the rank aggregation literature \cite{dwork_rankagg_2001,kenyon-mathieu_mallowsmle_2007,ailon_mallowsmle_2008}.
In this work we use a 
simple greedy algorithm as shown in Algorithm~\ref{alg:mallows_mle}.

\begin{algorithm}[t]
\caption{Computing MLE ranking for Mallows Model}
\label{alg:mallows_mle}
\begin{algorithmic}[1]
\STATE $C \leftarrow \assignments$ \hfill \COMMENT{$C$ contains unranked items}
\FOR{ $i = 1 \dots |\assignments|$}
\FOR{ $\assignment \in C$}
\STATE $\!\!x_\assignment \leftarrow \sum_{g \in G} \eta_g  |d'\in C:\!d'\succ_{\sigma_g}d|\!-\!|d'\in C: d\succ_{\sigma_g}d'|\!\!$\label{algline:mallows_score}
\ENDFOR
\STATE $\assignment^* \leftarrow \min_{d \in C} x_d$ \hfill \COMMENT{Select highest scoring item}
\STATE $r_{d^*}^{(\orderingest)} \leftarrow i$ \hfill \COMMENT{Rank as next item}
\STATE $C \leftarrow C / d^*$  \hfill \COMMENT{Remove $d^*$ from candidate set}
\ENDFOR
\STATE {\bf return} $\orderingest$
\end{algorithmic}
\end{algorithm}

\longversion{
In addition to this we also implement and empirically compare with a Borda count method which orders as per the average rank (leading to a 5-approximation for the case of full rankings \cite{fagin_mallowsmlepartial_2004}).
In other words, Line~\ref{algline:mallows_score} of Algorithm~\ref{alg:mallows_mle} is replaced with
$$x_d \leftarrow \sum_{g \in G} r_d^{(\sigma_g)}.$$
We denote this method as $\mathbf{MAL_{BC}}$.
A final variant of these techniques is one where {\em local Kemenization} is performed on the output ranking \ie adjacent pairs are swapped (in a bubble-sort like manner) if it increases the likelihood \cite{dwork_rankagg_2001}.
Variants which use local Kemenization are denoted with a $\mathbf{+K}$ suffix.
}
{
As an alternative algorithm for computing the estimated ranking, we utilize a Borda count-like approximation for the Mallows model (which we denote as MAL$_{BC}$), where  Line~\ref{algline:mallows_score} of Alg~\ref{alg:mallows_mle} is replaced with
$$x_d \leftarrow \sum_{\graderIngraders} r_d^{(\orderinggradespec)}.$$
}


\subsubsection{Score-Weighted Mallows (MALS)}
Mallow's model presented above has two shortcomings. First, it does not output a meaningful cardinal grade for the assignments, which makes it applicable only to ordinal grade estimation. Second, the distance $\delta_K$ does not distinguish between misordering assignments that are similar in quality from those that have a large quality difference.

To address these two shortcomings, we propose an extension which estimates cardinal grades $\cardinalest{d}$ for all assignments. To this effect, we introduce the following score-weighted ranking distance, which scales the distance induced by each misranked pairs by its estimated grade difference.
 \begin{defn} \label{defn:scoredeltakt}
 The {\bf score-weighted Kendall-$\tau$ distance} $\delta_{SK}$ over rankings $\sigma_1$, $\sigma_2$ given cardinal scores $\cardinal{d}$ is
 \begin{equation}
 \delta_{SK}(\sigma_1,\sigma_2|\cardinalvec )\!=\sum_{d_1 \succ_{\sigma_1} d_2} \!\!\!\!(\cardinal{d_1} - \cardinal{d_2}) \mathbb{I}[[d_2 \succ_{\sigma_2} d_1]]. \label{eq:scorekendalltau_delta}
\end{equation}
\end{defn}
Treating ties in the grader rankings as described above results in a score-weighted version of the Mallows model (MALS). We use the following maximum a posteriori estimator to estimate the scores $\cardinalestvec$.  
\begin{equation}
\cardinalestvec = \argmax_\cardinalvec \!\!\left\{ \!\! Pr(\mathbf{\cardinalvec})\!\prod_{g \in G}\!\frac{ \sum_{\ordering' \sim \orderinggradespec}\exp{(-\delta_{SK}(\orderingest,\ordering'|\cardinalvec))}  }{\!\!\! \sum\limits_{\ordering' \in \pi(D_g)}  \exp{(-\delta_{SK}(\orderingest,\ordering'|\cardinalvec))}    } \right\} \label{eq:swd_hinge_likelihood}  
\end{equation}
Note that $\orderingest$ can be obtained by sorting items as per $\cardinalest{d}$. $Pr(\mathbf{\cardinalestvec})=\prod_{d\in D}Pr(\cardinalest{d})$ is the prior on the latent item scores.
In our experiments we model $Pr(\cardinalest{d})\!\sim\!\mathcal{N}(0,9)$, and use the same prior in all of our methods.
While the resulting objective is not necessarily convex, we use SGD for grade estimation and initialize the grades using a scaled-down Mallows solution.

\subsubsection{Bradley-Terry Model (BT)}
\label{sec:logisticpairmodel}
The above models define distributions over rankings as a function of a ranking distance, and they require approximate methods for solving the maximum likelihood problem.
As an alternative, we can utilize rank aggregation models based on distributions over pairwise preferences, since a ranking of $n$ items can also be viewed as a set of preferences over the $\binom{n}{2}$ item pairs.
The Bradley-Terry model \cite{bradley_1952} is one model for pairwise preferences, and it derives a distribution based on the differences of underlying item scores $\cardinal{\assignment}$ through a logistic link function. 
\begin{eqnarray}
P(d_i \preferredto{\preferencesgradspec} d_j|\cardinalvec) = \frac{1}{1 + e^{-(\cardinal{d_i} - \cardinal{d_j})}}
\end{eqnarray}
Since each preference decision is modeled individually, the feedback from the user could be a (possibly inconsistent) set of preferences that does not necessarily have to form a consistent ordering. The following is the maximum a posteriori estimator used in this paper.
\begin{equation}
 \cardinalestvec = \argmax_\cardinalvec \left\{ Pr(\mathbf{\cardinalvec}) \prod_{g \in G} \:\prod_{d_i \preferredto{\preferencesgradspec} d_j } \frac{1}{1 + e^{-(\cardinal{d_i} - \cardinal{d_j})}} \right\} \label{eq:lp_likelihood}
\end{equation}
The resulting objective is (jointly) convex in all of the estimated grades $\cardinalest{d}$, with the gradients taking a simple form. Hence SGD can be used to estimate the grades efficiently and globally optimally.
\shortversion{
We treat ties as the absence of a preference. One can also extend this model to incorporate ties more explicitly, but we do not discuss this for  brevity.
}

\longversion{
If we are given cardinal scores and want to enforce a margin of some form then we use the following form (referred to as {\bf LPMM} - LPM with Margin):
\begin{equation}
 P(\rho_g) = \prod_{(d_i \succ d_j) \in \rho_g} \frac{1}{1 + e^{-\eta_g \cdot (s_i - s_j + y_{d_j}^{(g)} - y_{d_i}^{(g)}  )}}
\end{equation}
}
\longversion{
{

\bf Handling Ties} The above formulation ignores all pairs containing ties. We extend it to handle ties resulting in the {\bf LPT} model:
\begin{align}
  L(\mathbf{s_d}) &= Pr(\mathbf{s_d}) \prod_{g \in G} \Big( \prod_{d_i \succ_{\rho_g} d_j } \frac{e^{\eta \cdot (s_i - s_j)}}{e^{\eta \cdot (s_i - s_j)} + 1 + e^{\eta \cdot (s_j - s_i)} } \nnp
 & \times \prod_{d_i \approx_{\rho_g} d_j } \frac{1}{e^{\eta \cdot (s_i - s_j)} + 1 + e^{\eta \cdot (s_j - s_i)} } \Big) \label{eq:lpt_likelihood}
\end{align}
This objective tries to ensure that tied items do not have vastly different scores, while retaining convexity in the $\mathbf{s_d}$ variables.
}

\longversion{
Lastly there are two kinds of pairwise preferences that we can use given a ranking $\sigma_g$:
\begin{enumerate}
 \item {\bf All-Pairs} : More expressive but requires more computation.
 \item {\bf Consecutive-Pairs} : Simple but computationally very quick.
\end{enumerate}
}


\subsubsection{Thurstone Model (THUR)}
\label{sec:thurstonemodel}
An alternate to the logistic link function of the Bradley-Terry model is to utilize a normal distribution for the pairwise preferences.
Like the Bradley-Terry model, the resulting Thurstone model \cite{thurstone_1927} model can be understood as a random utility model using the following process:  For each pair of items $\assignment_i,\assignment_j$, the grader samples (latent) values $x_{\assignment_i}^{(g)} \sim \mathcal{N}(s_{\assignment_i},\frac{1}{2})$ and $x_{d_j}^{(g)} \sim \mathcal{N}(s_{\assignment_j},\frac{1}{2})$, and then orders the pair based on the two values. The mean of the normal distribution of $\assignment_i$ is the quality $\cardinal{\assignment_i}$. 
Maximum a posteriori estimation of the scores $\cardinalvec$ requires maximization of the following function:
\begin{equation}
  \cardinalestvec = \argmax_\cardinalvec \left\{ Pr(\mathbf{\cardinalvec}) \prod_{g \in G} \: \prod_{d_i \preferredto{\preferencesgradspec} d_j } \mathcal{F}(\cardinal{\assignment_i} - \cardinal{\assignment_j}) \right\} \label{eq:np_likelihood}
\end{equation}
$\mathcal{F}$ is the CDF of the standard normal distribution.
This objective function is convex and we use SGD to optimize it.

\longversion{
We can similarly enforce a margin here and either sets of pairwise preferences (All or Consecutive).
}

\longversion{
{\bf Handling Ties} We can also extend this model to account for ties $d_i \approx d_j$ by incorporating an additional parameter $\beta$ s.t. if $|x_{d_i}^{(g)} - x_{d_j}^{(g)}| \le \beta$ then we observe a tie. The resulting {\bf NPT} model has the following likelihood:
\begin{align}
 &L(\mathbf{s_d}) = Pr(\mathbf{s_d}) \prod_{g \in G} \prod_{d_i \succ_{\rho_g} d_j } \mathcal{F}(\sqrt{\eta}(s_{d_i} - s_{d_j} -\beta)) \times \nnp
 &\!\!\!\prod_{d_i \approx_{\rho_g} d_j }\!\!\!\Big(\mathcal{F}(\sqrt{\eta}(s_{d_i} - s_{d_j} + \beta))\!-\!\mathcal{F}(\sqrt{\eta}(s_{d_i} - s_{d_j} -\beta))\!\Big)  \label{eq:npt_likelihood}
\end{align}
In all our experiments we set $\beta=0.5$.
}


\subsubsection{Plackett-Luce Model (PL)}
\label{sec:plackettlucemodel}
A drawback of the pairwise preference models is that they can be less expressive than models built on distributions over rankings.
An extension to the Bradley-Terry model (the Plackett-Luce model \cite{plackett_1975}) allows us to use distributions over rankings, while still retaining convexity and simplicity of gradient computation.
This model can be best understood as a multi-stage experiment where at each stage, an item $d_i$ is drawn (w/o replacement) with probability $\propto e^{\cardinal{d_i}}$. 
Thus, the probability of observing ranking $\orderinggradespec$ under this process is:
$$P(\orderinggradespec|\cardinalvec) = \prod_{d_i \in D_g} e^{\cardinal{d_i}} / \Big( e^{\cardinal{d_i}} + \sum\limits_{d_i \preferredto{\orderinggradespec} d_j} e^{\cardinal{d_j}} \Big) $$
The resulting maximum a posteriori estimator is 
\begin{equation}
 \cardinalestvec = \argmax_\cardinalvec \left\{ Pr(\mathbf{\cardinalvec}) \prod_{g \in G} \: \prod_{d_i \in D_g} \frac{ e^{\cardinal{d_i}} }{ e^{\cardinal{d_i}} + \!\!\!\sum\limits_{d_i \preferredto{\orderinggradespec} d_j} e^{\cardinal{d_j}} } \right\}.  \label{eq:clr_likelihood}
\end{equation}

\longversion{
{\bf Handling Ties} A small modification to the likelihood from Eqn.~\ref{eq:clr_likelihood}, allows us to model ties:
\begin{equation}
 L(\mathbf{s_d}) = Pr(\mathbf{\cardinalestvec}) \prod_{g \in G} \prod_{d_i \in D_g} \frac{ e^{\eta_g s_{d_i}} }{ e^{\eta_g s_{d_i}}\! + \!\!\!\!\!\!\!\!\!\! \sum\limits_{d_j: d_i \succ_{\sigma_g} d_j}\!\!\!\!\!\!\!\! e^{\eta_g s_{d_j}} \! + \!\!\!\!\!\!\!\!\!\! \sum\limits_{d_j: d_i \approx_{\sigma_g} d_j} \!\!\!\! e^{\eta_g s_{d_j}} } \nn 
\end{equation}
The above {\bf CLRT} likelihood tries to ensure that documents tied to each other do not have scores that differ greatly, similar to the idea behind the LPT likelihood (Eqn.~\ref{eq:lpt_likelihood})
}


\longversion{

\subsection{Itemwise Rank Model (IR)}

}

\begin{algorithm}[t]
\caption{Alternating SGD-based Minimization}
\label{alg:overall_sgd}
\begin{algorithmic}[1]
\REQUIRE $N \ge 0$ (Number of iterations), Likelihood $L$
\STATE $Obj \leftarrow -\log{L}$
\STATE $\mathbf{\cardinalestvec} \!\leftarrow SGD_S(Obj, \mathbf{\graderreliabilityvec}\!=\!\mathbf{1})$ \hfill \COMMENT{Est. scores w/o reliabilities}
\FOR{ $i = 1 \dots N$}
\STATE $\mathbf{\graderreliabilityvec} \leftarrow SGD_G(Obj, \mathbf{\cardinalestvec})$ \hfill \COMMENT{Estimate reliabilities}
\STATE $\mathbf{\cardinalestvec} \leftarrow SGD_S(Obj, \mathbf{\graderreliabilityvec})$ \hfill \COMMENT{Est. scores with reliabilities}
\ENDFOR
\STATE {\bf return} $\mathbf{\cardinalestvec},\mathbf{\graderreliabilityvec}$
\end{algorithmic}
\end{algorithm}

\subsection{Grader Reliability Estimation} \label{sec:grader_reliability_estimation}
While the methods discussed in Section~\ref{sec:grade_estimation} allow us to estimate assignment grades from ordinal feedback, they still do not give us a means to directly estimate grader reliabilities $\graderreliabilityspecest$.
However, there is a generic way of extending all methods presented above to incorporate grader reliabilities. Using Mallow's model as an example, we can introduce $\graderreliabilityspecest$ as a variability parameter as follows:
\begin{eqnarray}
Pr(\ordering|\orderingmean,\graderreliabilityspec) = \frac{\sum\limits_{\ordering' \sim \orderinggradespec} \exp{(-\graderreliabilityspec\deltakerror(\orderingmean,\ordering'))}}{Z_{M}(\graderreliabilityspec,|\graderassignspec|)}
\end{eqnarray}
The resulting estimator of both $\orderingest$ and $\graderreliabilityvecest$ is
\begin{eqnarray}
 \orderingest,\graderreliabilityvecest \!= \!\argmax_{\ordering,\graderreliabilityvec} \!\!\left\{ \!\!\prod_{g \in G}\!\!Pr(\graderreliabilityspec)\frac{ \sum_{\ordering' \!\sim\! \orderinggradespec} \exp{(-\graderreliabilityspec \deltakerror(\!\ordering,\!\ordering'\!))}  }{Z_{M}(\graderreliabilityspec,|D_g|)} \!\!\right\},  \label{eq:mallows+Gmodel_likelihood}
\end{eqnarray}
where $Pr(\graderreliabilityspecest)$ is the prior on the grader reliability.
In this work we use a {\em Gamma} prior $\graderreliabilityspecest \sim Gamma(10,0.1)$.

Similarly, the other objectives can also be extended in this manner as seen in Table~\ref{tab:methodsummary}.
While many of the extended objectives, such as the one above in Eq.~(\ref{eq:mallows+Gmodel_likelihood}), are convex in the grader reliabilities $\graderreliabilityspecest$ (for given $\orderingest$), they unfortunately are not jointly convex in the reliabilities {\em and} the estimated grades.
We thus use an iterative alternating-maximization technique, which alternates between minimizing the objective to estimate the assignment grades and minimizing the objective to estimate the grader reliabilities.
This iterative alternating approach using stochastic gradient descent is used for all joint estimation tasks in this paper.
Note that methods which estimate the reliabilities using Algorithm~\ref{alg:overall_sgd} are denoted by a {\bf +G} suffix to the method, while those that simply estimate the assignment grades are represented by the method name alone.

\begin{table}[t]
\scriptsize
\addtolength{\tabcolsep}{-1.5mm}
\centering
\vspace{-0.001in}
 \begin{tabular}{|c|E{0.42in}c|c|p{2.23in}G|}
 \hline
Method & Ties? & Score & Cnvx & Estimator & Variants \\ \hline
MAL+G & No & No & No & $\!Pr(\mathbf{\graderreliabilityvec}) \prod_{\graderIngraders}\!\!\!\!\sum\limits_{\ordering' \sim \orderinggradespec} \!\!\!\exp{\!(-\!\graderreliabilityspecest\!\deltakerror(\!\orderingest,\!\ordering'\!))}\!/\!Z_{M}(\!\graderreliabilityspecest,\!|\graderassignspec|) $ & $MAL_{BC}$,$MAL+K$ \\
MALS+G & No & Yes & No & $\!Pr(\mathbf{\cardinalestvec},\!\mathbf{\graderreliabilityvec}) \prod_{\graderIngraders}\!\!\!\!\!\sum\limits_{\ordering' \sim \orderinggradespec}\!\!\!\!\!\exp{\!(-\graderreliabilityspecest \deltaskerror(\orderinggradespec,\orderingest,F))}  / Z(\cdot) $ &  \\
BT+G & No (LPT:Yes) & Yes & Yes & $\!Pr(\mathbf{\cardinalestvec},\!\mathbf{\graderreliabilityvec}) \prod_{\graderIngraders} \prod_{\assignment_i\!\preferredto{\!\preferencesgrad{\grader}}\!\assignment_j} \!\! 1/(1\!+\!e^{\!-\graderreliabilityspecest (\score{\!\assignment_i}\!-\!\score{\assignment_j})})\!$ & LPM \\
THUR+G & No (NPT:Yes) & Yes & Yes & $\!Pr(\mathbf{\cardinalestvec},\!\mathbf{\graderreliabilityvec}) \prod_{\graderIngraders} \prod_{\assignment_i\!\preferredto{\preferencesgrad{\grader}}\!\assignment_j} \mathcal{F}(\sqrt{\graderreliabilityspecest} (\score{\!\assignment_i}\!-\!\score{\assignment_j}) )$ & NPM \\
PL+G & No (CLT:Yes) & Yes & Yes & $\!Pr(\mathbf{\cardinalestvec},\!\mathbf{\graderreliabilityvec})\!\!\prod\limits_{\graderIngraders}\!\prod\limits_{\assignment_i \in \graderassignspec} \!\!\! 1 / (1 \!+ \!\!\!\!\!\!\!\sum\limits_{\assignment_i\preferredto{\preferencesgrad{\grader}}\!\assignment_j} \!\!\!\!\!\!\!e^{-\!\graderreliabilityspecest(\score{\!\assignment_i}\!-\!\score{\assignment_j}) } ) $ & CLRM\\
\hline
\end{tabular}
\vspace{-0.001in}
\caption{\label{tab:methodsummary} Summary of the ordinal methods studied which model the grader's reliabilities, including the  
ability to output cardinal scores and if the resulting objective is convex in these scores.}
\end{table}

\section{Experiments}

In the following we present experiments that compare ordinal and cardinal peer grading methods. We evaluate their ability to predict instructor grades, their variability, their robustness to bad peer grading, and their ability to identify bad graders. We also present the results from a qualitative student survey to evaluate how students perceived the peer grading process.

\subsection{Data Collection in Classroom Experiment}
\label{sec:datasets}

We use a real dataset consisting of peer feedback, TA grades, and instructor grades for evaluating the peer grading methods proposed in this paper. This data was collected as part of a senior-undergraduate and masters-level class with an enrollment of about 170 students. The class was staffed with 9 Teaching Assistants (TAs) 
that participated in grading, and a single Instructor. This size of class is attractive, since it is large enough for collecting a substantial number of peer grades, while at the same time allowing traditional instructor and TA grading to serve as a baseline. The availability of instructor grades makes our data different from other peer-grading evaluations used in the past (e.g., \cite{piech_peergrading_2013}). 
We plan to make the data available subject to IRB approval.

The dataset consists of two parts that were graded independently, namely the {\em poster presentation} and the {\em final report} of an 8-week long course project. Students worked in groups of 3-4 students for the duration of the project, and there were a total of 44 project groups.
While student worked in groups, peer grading was performed individually via the Microsoft Conference Management Toolkit (CMT) system.
The peer grading process was performed double-blind, and the reviewer assignments were made randomly.
Students were given clear directives and asked to focus on aspects such as {\em novelty} and {\em clarity} (among others) while determining their grade. They were also asked to justify their grade by providing feedback comments. Students were told that a part of their grade depends on the quality of their peer feedback. 

All grading was done on a 10-point (cardinal) Likert scale, where 10 was labeled ``perfect'', 8 ``good'', 5 ``borderline'', 3 ``deficient'' and 1 ``unsatisfactory''. This will allow us to compare cardinal and ordinal peer grading methods, where ordinal methods merely use the ordering (possibly with ties) implied by the cardinal scores. Note that in a true application of ordinal peer grading accuracy could improve, since it would allow simplifying the grading instructions and reduce cognitive overhead if students did not have to worry about the precise meaning of specific cardinal grades.
 
The following describes the grading processes used at each stage, and Table~\ref{tab:datasetstats} summarizes some of the key statistics.

\begin{table}[t]
\scriptsize
\addtolength{\tabcolsep}{-1.5mm}
\centering
 \begin{tabular}{|c|Gc|c|}
 \hline
 Data Statistic & PR & PO & FR \\ \hline
Number of Assignments & 44 & 42 & 44 \\
Number of Peer Reviewers & 158 & 148 & 153 \\
Total Peer Reviews & 614 & 996 & 586 \\
Total TA Reviews & 44 & 78 & 88  \\
Participating TAs & 5 & 7 & 9  \\
\longversion{ No. of Grade levels  (1 to ?) & 3 & 10 & 10 \\ }
Per-Item Peer Grade Devn. & 0.58 & 1.16 & 1.03 \\
\hline
\end{tabular}
 \begin{tabular}{|c|c|c|cG|}
 \hline
 Set & Who? & Mean & Devn. & Skew  \\ \hline
 \longversion{
 \multirow{2}{*}{PR} & Peers & 2.30 & 0.67 & -0.42\\ 
  & Staff & 1.82 & 0.53 & -0.13\\ \hline
  }
 \multirow{3}{*}{PO} & Peers & 8.16 & 1.31 & -0.63 \\ 
  & TAs & 7.46 & 1.41 & -2.03\\
  & Meta & 7.55 & 1.53 & -2.83\\ \hline
 \multirow{3}{*}{FR} & Peers & 8.20 & 1.35 & -1.17\\ 
  & TAs & 7.59 & 1.30 & -0.69\\
  & Instructor & 7.43 & 1.16 & -0.45\\ \hline
\end{tabular}
\vspace{-0.01in}
\caption{\label{tab:datasetstats} Statistics for the two datasets (PO=Poster, FR=Report) from the classroom experiment along with the staff (TAs/Meta/Instructor) and student grade distributions.}
\vspace{-0.001in}
\end{table}

\subsubsection{Grading Process for Poster Presentations}

The poster presentations took place in a two-hour poster session. Two groups did not present their poster. Students were encouraged to rotate presenting their poster. This likely increased variability of grades, since different reviewers often saw different presenters. Students and TAs took notes and entered their reviews via CMT afterwards.

The {\em TA Grades} were independent, meaning that the TAs did not see the peer reviews before entering their review. There were on average 1.85 TA reviews for each poster.

The {\em Peer Grades} totaled on average 23.71 reviews for each poster, with each peer reviewer reviewing 6.73 posters on average.

The final {\em Meta Grade} for each poster was determined as follows. One of the TAs that already provided an independent review was selected as a meta-reviewer. This TA was asked to aggregate all the arguments brought forward in the reviews and make a final grade on the same 10-point scale. The instructor oversaw this process, but intervened only on very few grades.

\subsubsection{Grading Process for Final Projects}

At the end of the project, groups submitted a report of about 10 pages in length. The reviewing process was similar to that of the poster presentations, but with one important difference --- namely that all project reports were graded by the TAs and the instructor without any knowledge of the peer reviews, as detailed below.

On average each report received 13.32 {\em Peer Grades} as the overall score on each of the peer reviews (students were also asked for component scores like ``clarity'', etc.).

Each report also received two {\em TA Grades}, which the TAs submitted without knowledge of the peer reviews.

Finally, each report received an {\em Instructor Grade}, following the traditional process of how projects were graded in this class. The instructor and head TA each graded half the projects and determined the grade based on their own reading of the paper, taking the TA reviews as input. These grades were provided without viewing the peer reviews. We can therefore view the instructor grades as an assessment that is entirely independent of the peer grades (in contrast to the Meta Grades for the posters, which have some dependency).

\subsection{Evaluation Metrics}
\label{sec:metrics}

A commonly used measure for reporting student performance (among many standardized tests) is the percentile rank relative to all students in the class.
Following this practice, we use percentile rank as the grade itself (a letter grade can easily be derived via curving), and report ranking metrics as our main indicators of performance.
In particular, we use the following variant of Kendall-$\tau$ that accounts for ties.
 \begin{align}
 &\deltakterror(\sigma_1,\sigma_2)\!=\!\!\!\!\sum_{d_1 \succ_{\sigma_1} d_2} \!\!\!\!\mathbb{I}[[d_2 \succ_{\sigma_2} d_1]] + \frac{1}{2} \mathbb{I}[[d_1 \approx_{\sigma_2} d_2]]  \label{eq:kendalltauties_delta} 
 \end{align}
Note that this measure is not symmetric, assuming that the first argument is a target ranking and the second argument is a predicted ranking. It treats ties in the target ranking as {\em indifference}. Ties in the predicted ranking are treating as a lack of information, incurring a $\frac{1}{2}$ error (\ie equivalent to breaking ties randomly).
 Such a correction is necessary for evaluation purposes, since otherwise predicted rankings with all ties (which convey no information) would incur no error.
Normalizing $\deltakterror(\sigma_1,\sigma_2)$ and accounting for the fact that we may have more than one target ranking leads to the following error measure.
\begin{defn} \label{defn:kterrors}
 Given a set of target rankings $S_g$, we define the {\bf Kendall-$\tau$ error} $\mathcal{E}_{K}$ 
 of predicted ranking $\sigma_I$ as:
 \begin{align}
  & \mathcal{E}_{K}(\sigma_I) = \frac{100}{|S_g|}  \sum_{\sigma_t \in S_{g}} \frac{\deltakterror(\sigma_t,\sigma_I)}{\max_{\sigma \in \pi(D)} \deltakterror(\sigma_t,\sigma) } \label{eq:kterr_all}
 \end{align}
\end{defn}
 This error {\em macro-averages} the (normalized) $\deltakterror$ errors for each target ranking.
 Due to the normalization, they lie between 0 (indicating perfect agreement) and 100\% (indicating reversal with target rankings). A random ranking has expected $\mathcal{E}_{K}$ error of 50\%.
 
 \longversion{
 
 Micro-averaged measures, as well as those w/o ties
 
 }

\begin{figure}[t]
\centering
\includegraphics[scale=0.332]{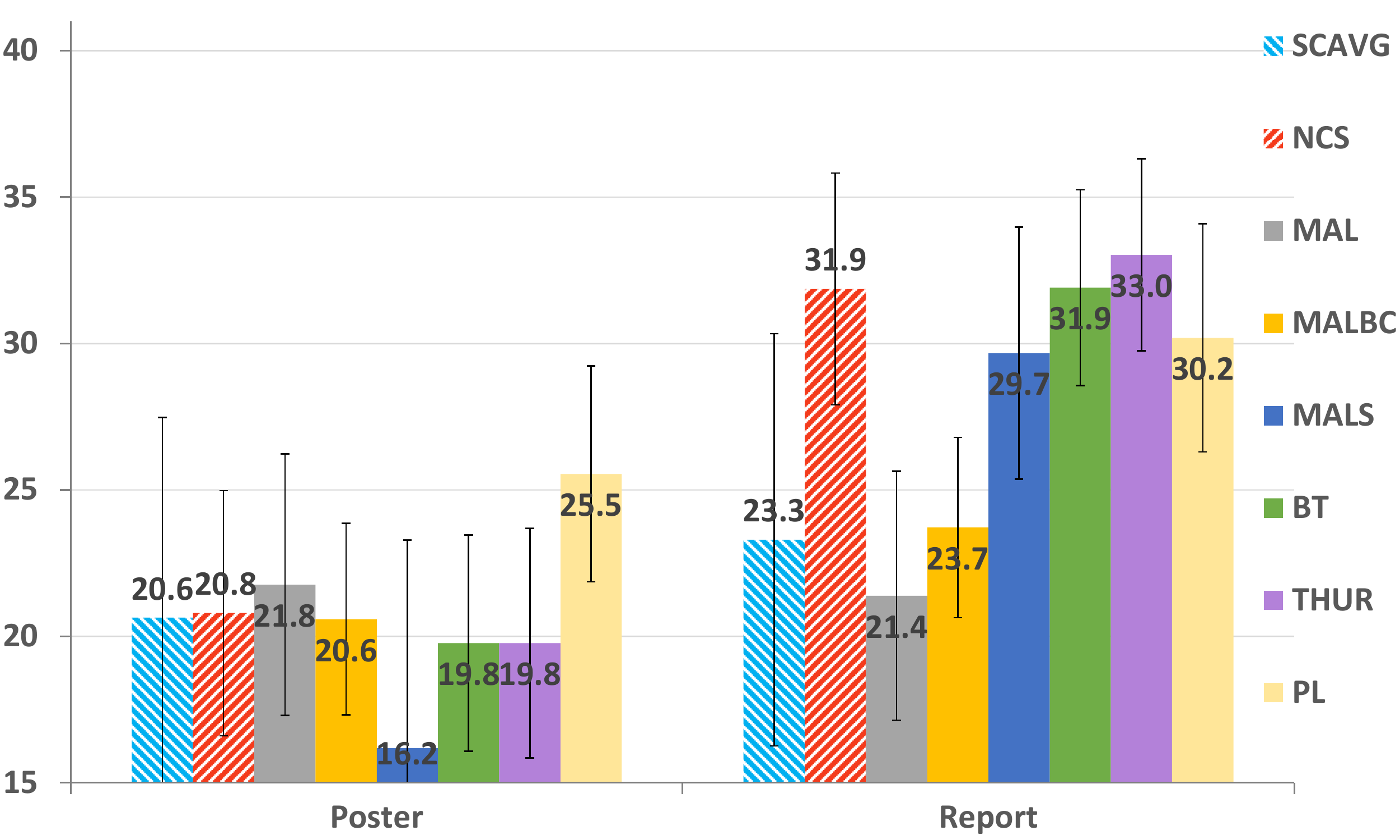}
\vspace{-0.2in}
\caption{Comparing peer grading methods (w/o grader reliability estimation) against Meta and Instructor Grades in terms of $\tauerror$ (lower is better). \label{fig:ordvscard_nog_barchart_instruc}}
\end{figure}

\subsection{How well does Ordinal vs. Cardinal Peer Grading Predict Final Grade?} \label{sec:comp_instr}

The first question we address is in how far peer grading resembles the grades given by an instructor. Specifically, we investigate whether ordinal peer grading methods achieve similar performance as cardinal peer grading methods, even though ordinal methods receive strictly less information.

For all methods considered in this paper, Figure~\ref{fig:ordvscard_nog_barchart_instruc} shows the Kendall-$\tau$ error $\tauerror$ compared to the Meta Grades for the Posters, and compared to the Instructor Grades for the Reports. The errorbars show estimated standard deviation using bootstrap-type resampling.

On the posters, none of the methods show significantly worse performance than another method. In particular, there is no evidence that the cardinal methods are performing better than the ordinal methods. A similar conclusion also holds for the reports. However, here the ordinal methods based on Mallow's model perform better than the cardinal NCS\footnote{We tuned the hyperparameters of the NCS model to maximize performance. We also used a fixed grader reliability parameter in the NCS model, since it provided better performance than with reliability estimation (NCS+G).} method \cite{piech_peergrading_2013} (see Algorithm~\ref{alg:ncs_baseline}), as well as some of the other ordinal methods. Simply averaging the cardinal scores of the peer graders, which we call Score Averaging (SCAVG), performs surprisingly well.

In summary, most methods achieve an $\tauerror$ between 20\% and 30\% on both problems, but all have large standard deviations. The $\tauerror$ appears lower for the posters than for the projects, which can be explained by the fact that the Meta Grade was influenced by the peer grades. But how good is an $\tauerror$ between 20\% and 30\%?

\begin{figure}[t]
\centering
\vspace{-0.04in}
\includegraphics[scale=0.332]{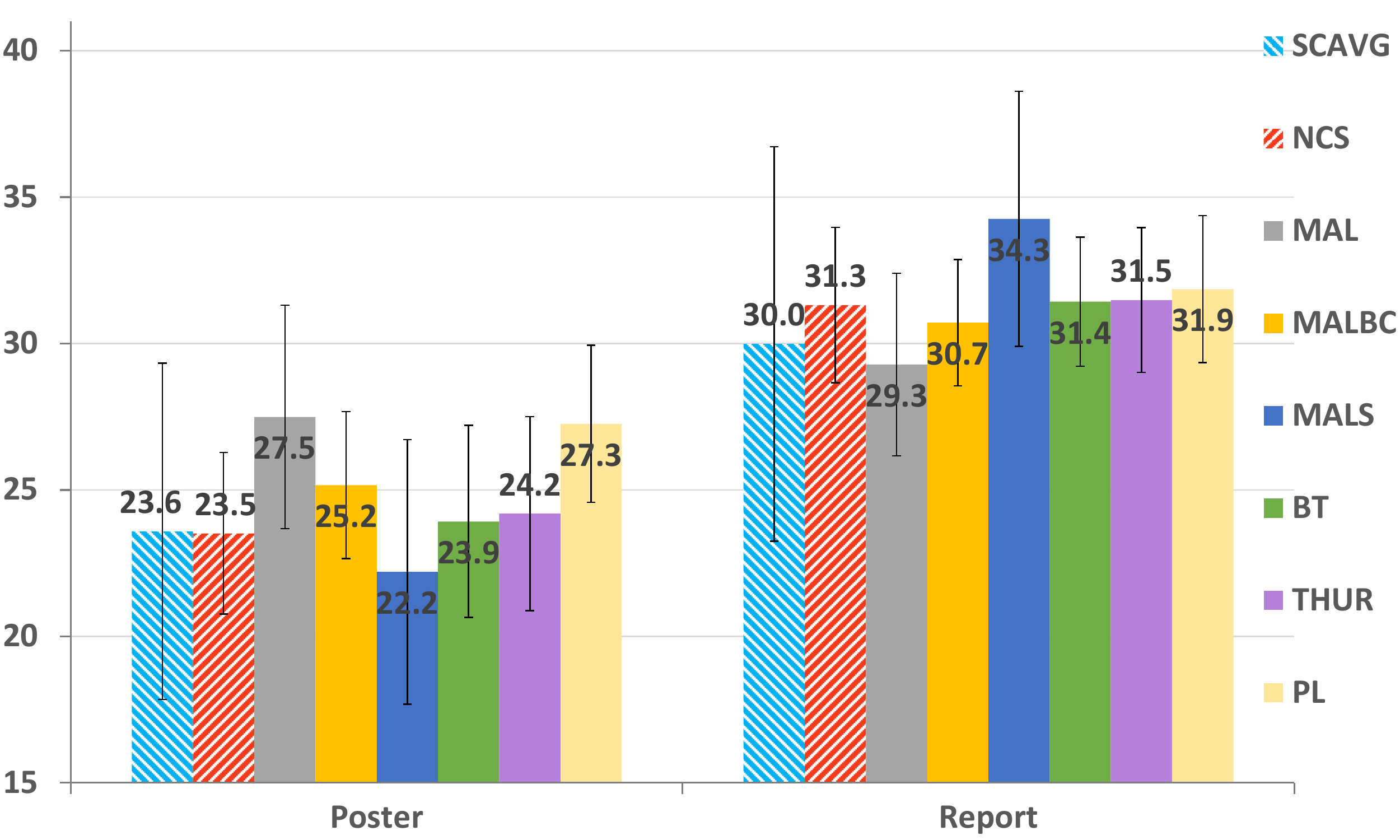}
\vspace{-0.25in}
\caption{Comparing peer grading methods (w/o grader reliability estimation) against TA Grades in terms of $\tauerror$, using TA grades as the target ranking. \label{fig:ordvscard_nog_barchart_ta}}
\vspace{-0.1in}
\end{figure}

\subsection{How does Peer Grading Compare to TA Grading?} \label{sec:comp_ta}

We now consider how Peer Grading compares to having each assignment graded by a TA. For medium sized classes, TA grading may still be feasible. It is therefore interesting to know if TA grading is clearly preferable to Peer Grading when it is feasible. But more importantly, the inter-judge agreement between multiple TAs can give us reference points for the accuracy of Peer Grading.

As a first reference point, we estimate how well the TA Grades reflect the Meta Grades for the posters and the Instructor Grades for the reports. In particular, we consider a grading process where each assignment is graded by a single TA that assigns a cardinal grade. Each TA grades a fraction of the assignments, and a final ranking of the assignments is then computed by sorting all cardinal grades. We call this grading process {\em TA Grading}.

We can estimate the $\tauerror$ of TA grading with the Meta Grades and the Instructor Grades, since we have multiple TA grades for most assignments. We randomly resample a TA grade from the available grades for each assignment, compute the ranking, and then estimate mean and standard deviation of the $\tauerror$ over 5000 samples. This leads to a mean $\tauerror$ of $22.0 \pm 16.0$ for the posters and $22.2 \pm 6.8$ for the reports. Comparing these to the $\tauerror$ of the peer grading methods in Figure~\ref{fig:ordvscard_nog_barchart_instruc}, we see that they are comparable to the performance of many peer grading methods --- {\em even though the $\tauerror$ of TA grading is favorably biased. Note that Meta Grades and the Instructor Grades were assigned based on the same TA grades we are evaluating against.} 

To avoid this bias and provide a fairer comparison with TA grading, we also investigated how consistent peer grades are with the TA grades, and how consistent TA grades are between different TAs. Figure~\ref{fig:ordvscard_nog_barchart_ta} shows the $\tauerror$ of the peer grading methods when using TA Grades as the target ranking for both the Posters and the Reports. Variances were again estimated via bootstrap resampling. Note that TA Grades were submitted without knowledge of the Peer Grades. Overall, the peer grades have an $\tauerror$ with the TA Grades that is similar to the $\tauerror$ with the respective Final grades considered in the previous subsection. Again, there is no evidence that the ordinal peer grading methods are less predictive of the TA Grades than the cardinal peer grading methods.

To estimate $\tauerror$ between different TAs, we use the following resampling procedure. In a leave-one-out fashion, we treat the grades of a randomly selected TA as the target ranking and compute the predicted ranking by sampling from the other TAs grades as described above. Averaging over 5000 repetitions reveals that the $\tauerror$ between the TAs is $47.5 \pm 21.0$ for the posters and $34.0 \pm 13.8$ for the reports. 

These numbers can be compared to the $\tauerror$ of peer grading methods in Figure~\ref{fig:ordvscard_nog_barchart_ta}. For the Reports, peer grades are roughly as consistent with the TA grades as other TA grades are. For the posters the peer grading methods are substantially more predictive of TA grades than other TA grades. The reason for this is at least twofold. First, the peer grading methods have access to much more data, which reduces variability (especially since presentations were not always given by the same student). Second, the peer grading methods have enough data to correct for different grading scales, while offsets in grading scales can have disastrous consequences in TA grading.

\begin{figure}[t]
\centering
\vspace{-0.04in}
\includegraphics[scale=0.332]{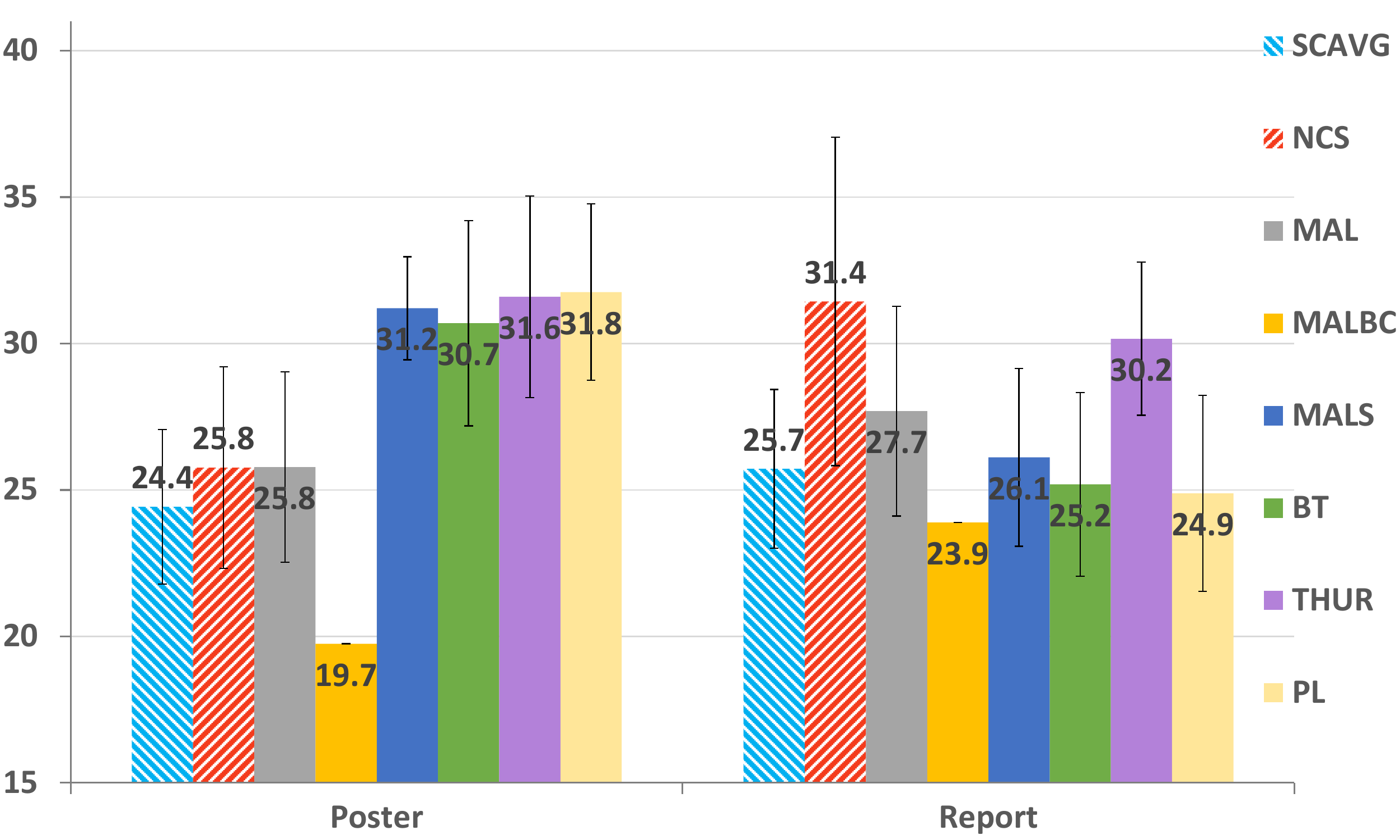}
\vspace{-0.25in}
\caption{Self-consistency of peer-grading methods (w/o grader reliability estimation) in terms of $\tauerror$.\label{fig:self-consistent}}
\vspace{-0.02in}
\end{figure}

Finally, we also consider the self-consistency of the peer grading methods. Analogous to the self-consistency of TA grading, we ask how similar are the grades we get if we repeat the grading procedure with a different sample of assessments. We randomly partition peer reviewers into two equally sized datasets. For each peer grading method, we perform grade estimation on both datasets, which generates two rankings of the assignments. Ties in these rankings are broken randomly to get total orderings. Figure~\ref{fig:self-consistent} 
shows the $\tauerror$ between the two rankings  (over 20 sampled partitions). For the posters, peer grading is substantially more self consistent than TA grading, and for the reports all peer grading methods have lower $\tauerror$ estimates than TA grading as well. 


Overall, we conclude that there is no evidence that TA grading would have led to more accurate grading outcomes than peer grading.

\begin{figure*}[!t]
\centering
\includegraphics[scale=0.332]{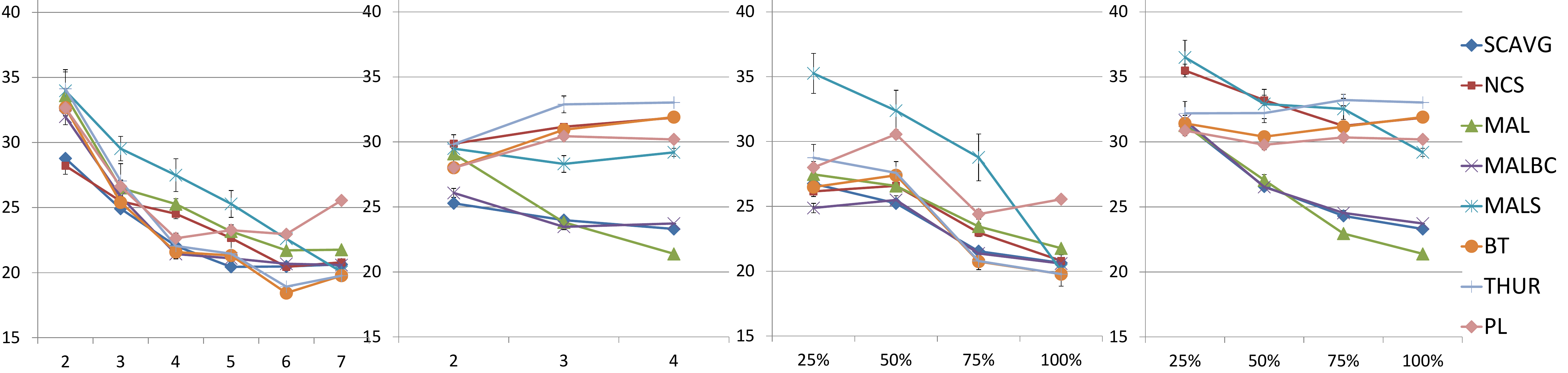}
\vspace{-0.2in}
\caption{Change in $\tauerror$ performance of peer grading methods (using Meta and Instructor Grades as target ranking) when varying the number of assignments assigned to each reviewer for Posters (first from left) \& Reports (second), and when varying the number of peer reviewers for Posters (third) Reports (last). \label{fig:scalereverror}}
\vspace{-0.01in}
\end{figure*}


\subsection{How does Grading Accuracy Scale with the Number of Peer Reviews?}
\label{sec:exp_scale}
How many reviewers are necessary for accurate peer grading, and how many reviews does each peer grader need to do? 
To gage how performance changes with the number of peer reviews, we performed two sets of experiments. 
First, we created 20 smaller datasets by downsampling the number of peer reviewers.
The results are shown in the two rightmost graphs of Figure~\ref{fig:scalereverror}.
Overall, the methods degrade gracefully when the number of reviewers is reduced. Overall, we find that most ordinal methods scale as well as cardinal methods for both datasets.

A second way of increasing or reducing the amount of available data lies in the number of assignments that each student grades.
Thus we repeated the experiment, but instead downsampled the number of assignments per reviewer (corresponding to a lower workload for each grader).
The leftmost two plots of Figure~\ref{fig:scalereverror} show the results.
Again, we find that performance degrades gracefully.

\begin{figure}[!t]
\centering
 \includegraphics[scale=0.332]{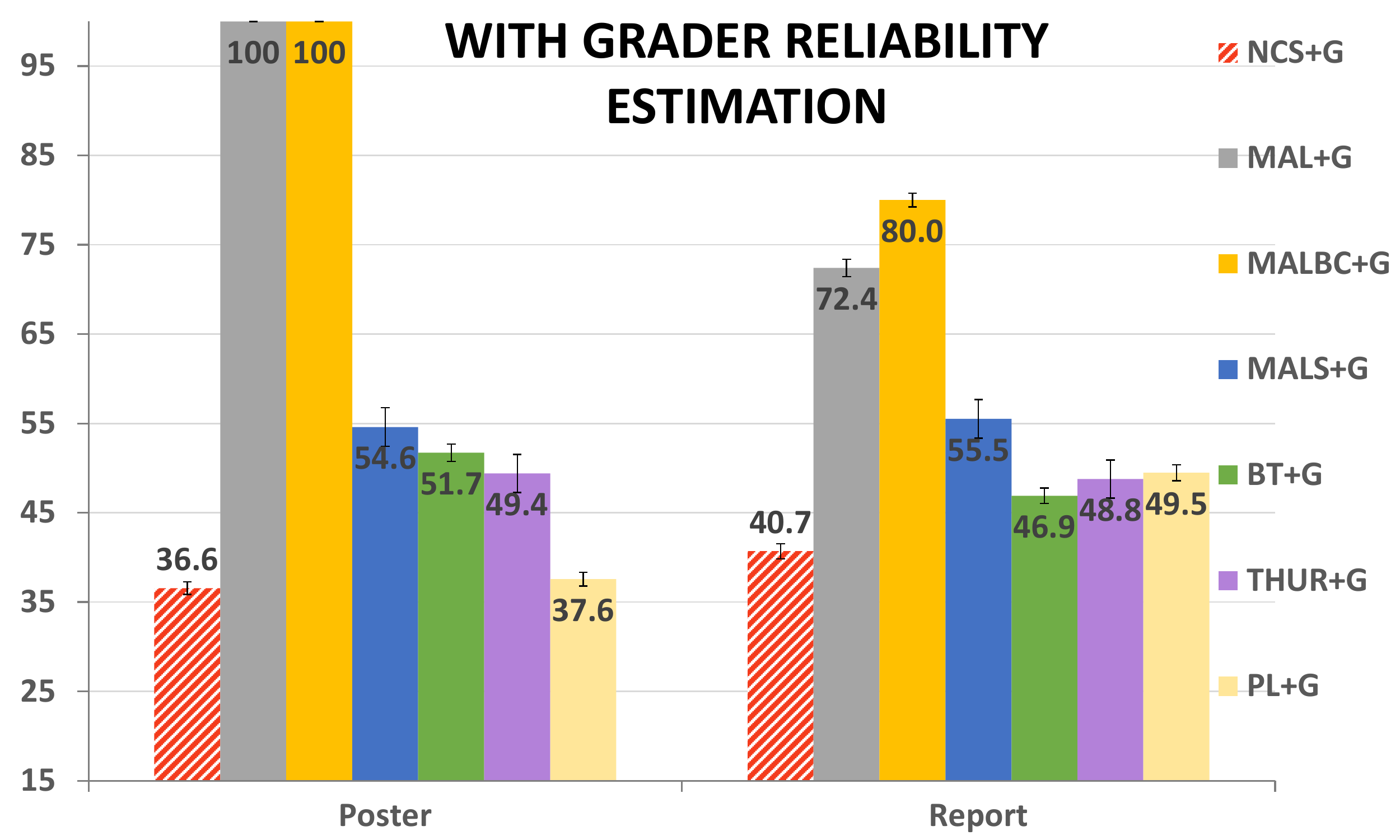}
 \includegraphics[scale=0.332]{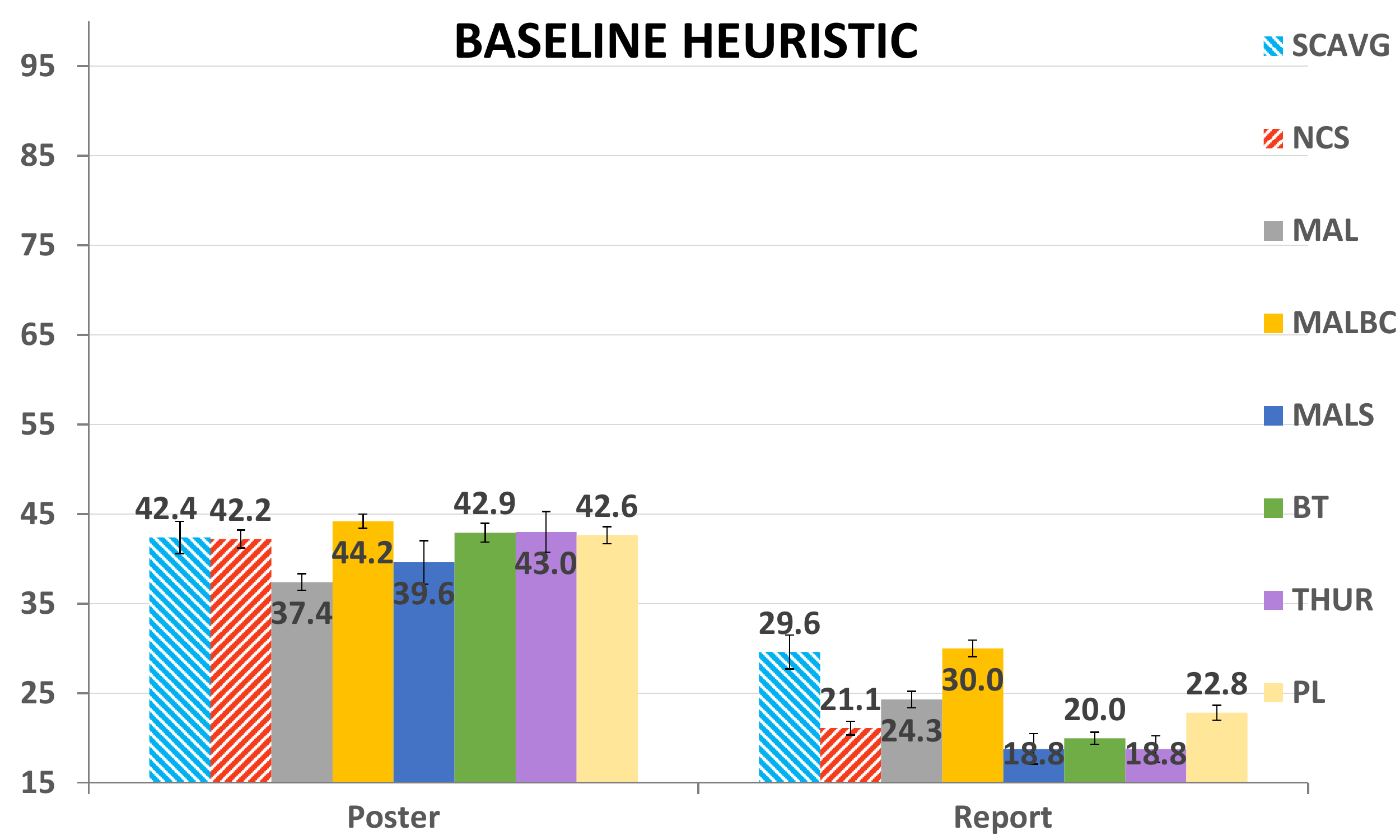}
\vspace{-0.2in}
\caption{Percentage of times an user who randomly scores and orders assignments is among the 20 least reliable graders (\ie bottom 12.5\%). \label{fig:uselrel-identifyrand}}
\end{figure}

\subsection{Can the Peer Grading Methods Identify Unreliable Graders?}
\label{sec:exp_reliability_howwell}


Peer grading can only work in practice, if graders are sufficiently incentivised to report an accurate assessment. This can be achieved by giving a grade also for the quality of the grading. In the following, we investigate whether the grader reliability estimators proposed in Section~\ref{sec:grader_reliability_estimation} can identify graders that are not diligent.

For both the posters and the projects, we add 10 ``lazy'' peer graders that report grades drawn from a normal distribution whose mean and variance matches that of the rest of the graders\footnote{Otherwise it would be easy to identify these graders.}.
For the ordinal methods, this results in a random ordering. We then apply the peer grading methods, estimating the respective reliability parameters $\eta_g$ for each grader using 10 iterations
of the alternating optimization algorithm. We then rank graders by their estimated $\eta_g$.

Figure~\ref{fig:uselrel-identifyrand} (top) shows the percentage of lazy graders that rank among the 20 graders with the lowest $\eta_g$. The error bars show standard error over 50 repeated runs with different lazy graders sampled. Most ordinal methods significantly outperform the cardinal NCS method for both the posters and the reports. The variants of Mallow's model perform very well, identifying around 70-80\% of the lazy graders for the reports and all 10 lazy graders for the posters. 
The better performance for the posters than for the reports was to be expected, since students provide 7 instead of 4 grades.

Figure~\ref{fig:uselrel-identifyrand} (bottom) shows the results of a heuristic baseline. Here, grade estimation without reliability estimation is performed, and then graders are ranked by their $\tauerror$ with the estimated ranking $\orderingest$.
For almost all methods, this performs worse, clearly indicating that reliability estimation is superior in identifying lazy graders.
We find similar results even when there are 100+ lazy graders, and we investigate robustness in the following section.

%

 \begin{figure}[!t]
\centering
\vspace{-0.02in}
\includegraphics[scale=0.332]{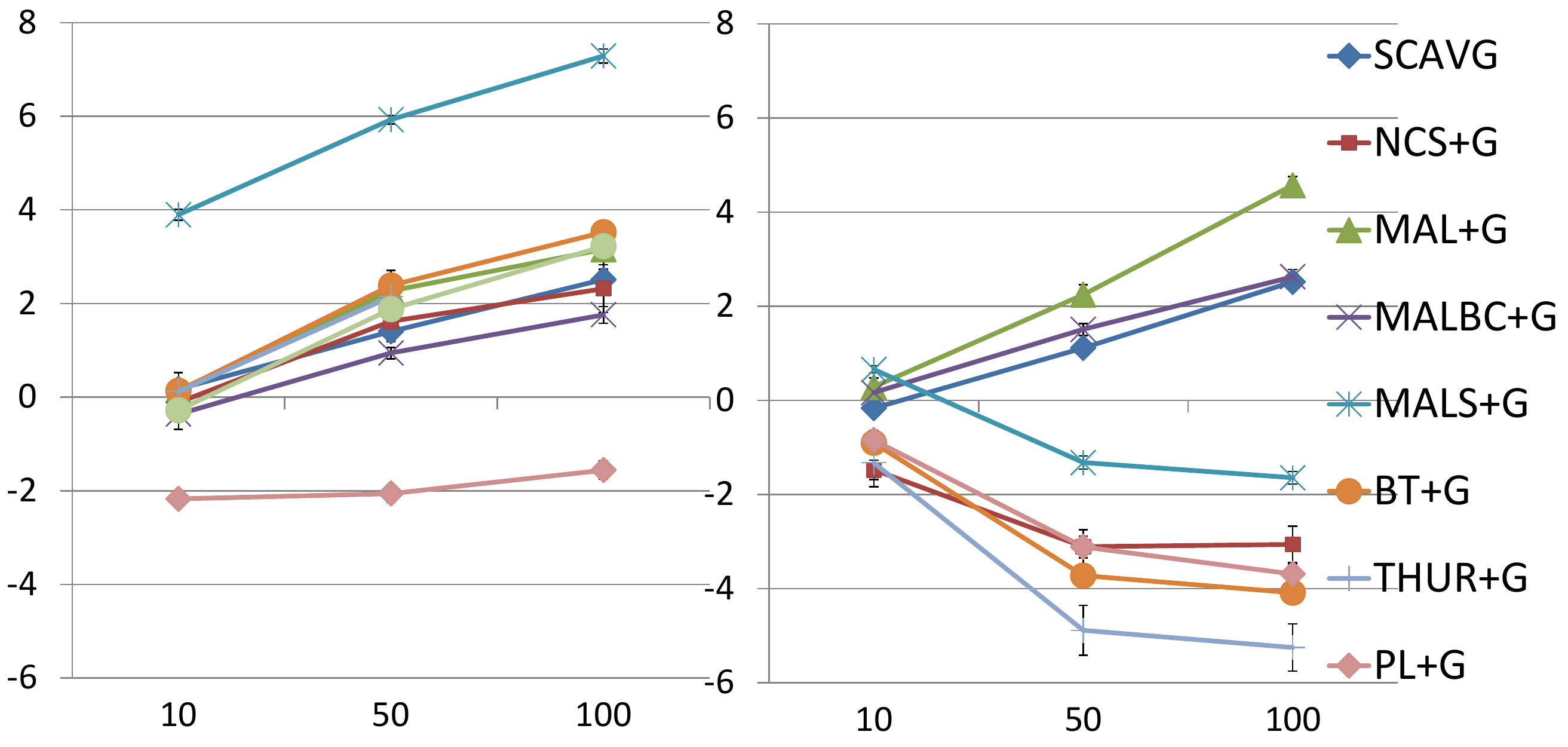}
\vspace{-0.22in}
\caption{Change in $\tauerror$ (using Instructor and Meta Grades as target ranking) for (Left) Posters and (Right) Final Reports with the addition of an increasing number of {\em lazy} graders \ie $\tauerror($With Lazy$) - \tauerror($Without Lazy$)$. A negative value indicates that performance improves on adding this noise. \label{fig:robustness-to=ffkrand-withg}}
\vspace{-0.01in}
\end{figure}

\subsection{How Robust are the Peer Grading Methods to Lazy Graders?}
\label{sec:exp_reliability_help}


While Section~\ref{sec:exp_reliability_howwell} showed that reliability estimation in ordinal peer grading is well-suited for identifying lazy graders, we would also like to know what effect these lazy graders have on grade estimation performance.
We study the robustness of the peer grading methods by adding an increasing number of lazy graders.
Figure~\ref{fig:robustness-to=ffkrand-withg} shows the change in $\tauerror$ (w.r.t. Instructor/Meta grades) after adding 10/50/100 lazy graders (compared to the $\tauerror$ with no lazy graders).
We find that in most cases performance does not change much relative to the variability of the methods.
Interestingly, in some cases performance also improves on adding this noise. 
A deeper inspection reveals that noise is most beneficial for methods whose original $\tauerror$ performance was weaker than that of the other methods.
For example, the Thurstone model showed the weakest performance on the Reports and improves the most.




\begin{table}[t]
\scriptsize
\addtolength{\tabcolsep}{-1.25mm}
\centering
\vspace{-0.001in}
 \begin{tabular}{|c|c|c|c|c|}
 \hline
   Method & \multicolumn{2}{c|}{Posters} & \multicolumn{2}{c|}{Reports} \\ \hline
   & Runtime & Runtime (+G) & Runtime & Runtime (+G) \\ \hline
NCS & 0.32 $\pm 0.03$  & 7.0 $\pm 0.55$  & 0.20 $\pm 0.03$  & 4.6 $\pm 0.25$ \\
MAL & 0.01 $\pm 0.00$  & 6.1 $\pm 0.11$  & 0.01 $\pm 0.00$  & 2.5 $\pm 0.03$ \\
MAL$_{BC}$ & 0.01 $\pm 0.00$  & 5.1 $\pm 0.08$  & 0.01 $\pm 0.00$  & 2.5 $\pm 0.03$ \\
MALS & 151.4 $\pm 12.39$  & 418.7 $\pm 9.10$  & 2.0 $\pm 0.13$  & 4.2 $\pm 0.16$ \\
BT & 0.46 $\pm 0.06$  & 5.6 $\pm 0.38$  & 0.21 $\pm 0.02$  & 2.2 $\pm 0.10$ \\
THUR & 57.9 $\pm 0.76$  & 490.1 $\pm 7.45$  & 12.2 $\pm 0.86$  & 120.8 $\pm 1.03$ \\
PL & 0.36 $\pm 0.03$  & 4.2 $\pm 0.08$  & 0.18 $\pm 0.01$  & 2.0 $\pm 0.10$ \\
\hline
\end{tabular}
\vspace{-0.001in}
\caption{\label{tab:runtime} Average runtime (with and without grader reliability estimation) and their standard deviation of different methods in CPU seconds.}
\vspace{-0.001in}
\end{table}

\subsection{How Computationally Efficient are  the Peer Grading Methods?}


While prediction accuracy is the prime concern of grade inference, computational efficiency needs to be sufficient as well. Table~\ref{tab:runtime} show the average runtimes and their standard deviations for the posters and the reports. All methods are tractable and most finish within seconds. 
The Score-Weighted Mallows model is less efficient for problems where a each grader assesses many assignments, since the gradient computations involves computing the normalization constant (which involves summing over all rankings). 
However, training scales linearly with the number of graders. 
Another method that requires more time is the Thurstone model.
The main bottleneck here is the computation of the gradient as it involves a lookup of a CDF value from the normal distribution table.


\subsection{Do Students Value Peer Grading?}

A final point that we would like to explore is that peer grading is not only about grade estimation, but also about generating useful feedback. In particular, the cardinal or ordinal assessments were only a small part of the peer feedback. Peer graders had to write a justification for their assessment and comment on the work more generally.

To assess this aspect of peer grading, a survey was conducted at the end of class as part of the course feedback process. 
This survey included two questions about the student's peer grading experience in the class; more specifically, about how {\em helpful} the feedback they received was, and how {\em valuable} the experience of providing feedback was to them.
Both questions were to be answered in free-form text. Of the 161 students that participated in the project, 120 students responded to at least one of the questions, with 119 answering the question about receiving feedback (mean response length in characters: 62.93; stdev. : 77.22) and 118 the question about providing feedback (mean: 100.36; stdev. : 105.74).
Following standard practice from survey analysis, we created five categories for coding these open-ended responses as show in Table \ref{tab:surveyquestions}. 
While the first four categories (roughly) follow a decreasing scale of approval, the last serves as a catch-all (including missing responses).

All free-text responses were manually assigned to these categories by four external annotators (who were not involved with the class and had not seen the comments before).
For all the 237 student comments (\ie responses), the annotators were asked to choose the category that was {\em most appropriate/best describes the comment}.
To check inter-annotator agreement we used the Fleiss Kappa measure.
$\kappa$ values of $0.8389$ and $0.6493$ for the two questions indicate high annotator agreement. 
The final assignment of response to category was done by majority vote among the four annotators (score of 0.5 each if tied between categories).


Table \ref{tab:surveyresults} summarizes the results of the survey after coding. Overall, around 68\% found it at least somewhat helpful to receive peer feedback, and around 74\% found substantial value in providing the peer feedback. Interestingly, of the 26\% of the students who expressed that receiving peer feedback was not (really) helpful to them, 17\% still found it valuable to provide peer feedback. 
Overall, we conclude that the vast majority of students found some value in the peer grading process.

\section{Conclusions}

\begin{table}[t]
\scriptsize
\addtolength{\tabcolsep}{-1.5mm}
\centering
\vspace{-0.001in}
 \begin{tabular}{|p{0.14in}|p{1.4in}||p{0.14in}|p{1.6in}|}
 \hline
\multicolumn{2}{|p{1.52in}||}{{\bf Question A}) Was getting peer feedback helpful?} & \multicolumn{2}{p{1.52in}|}{{\bf Question B}) Was providing peer feedback valuable?} \\ \hline
$A_1$ &  Yes, it was helpful. & $B_1$ & Yes it was a valuable experience \\
$A_2$ & Helpful, but not as much as instructor feedback. & $B_2$ & Yes, it was valuable, but with caveats (e.g. took lot of time). \\
$A_3$ & Somewhat helpful (e.g. only few comments were helpful). & $B_3$ & Only little value (e.g. was too difficult / lacked the grading skills) \\
$A_4$ & No / Not really / Did not help much. & $B_4$ & Not valuable / Not really valuable. \\
$A_5$ & Other / Missing & $B_5$ & Other / Missing\\ \hline
\end{tabular}
\vspace{-0.1in}
\caption{\label{tab:surveyquestions} Response categories for survey questions.}
\vspace{-0.001in}
\end{table}

\begin{table}[t]
\scriptsize
\addtolength{\tabcolsep}{-0.75mm}
\centering
 \begin{tabular}{|c|c|c|c|c|c||c|}
 \hline
 \% & $A_1$ & $A_2$ & $A_3$  & $A_4$  & $A_5$ & Total \\ \thickhline
$B_1$ & \cellcolor{gray!30}{\bf 34.58} & 2.08 & \cellcolor{gray!5}5.83 & \cellcolor{gray!10}10.00 & 1.67 & \cellcolor{gray!50}{\bf 54.17} \\ \hline
$B_2$ & \cellcolor{gray!5}5.42 & 0.00 & \cellcolor{gray!5}5.83 & \cellcolor{gray!5}7.08 & 1.67 & \cellcolor{gray!20}20.00 \\ \hline
$B_3$ & 0.42 & 2.92 & 2.08 & 2.50 & 0.42 & \cellcolor{gray!5}8.33 \\ \hline
$B_4$ & 2.92 & 0.83 & \cellcolor{gray!5}5.00 & \cellcolor{gray!5}5.42 & 0.00 & \cellcolor{gray!10}14.17 \\ \hline
$B_5$ & 0.00 & 0.00 & 0.42 & 1.67 & 1.25 & 3.33 \\ \hline
\hline
 Total & \cellcolor{gray!40}{\bf 43.33} & \cellcolor{gray!5}5.83 & \cellcolor{gray!15}19.17 & \cellcolor{gray!25}26.67 & \cellcolor{gray!5}5.00 & \\ \hline
\end{tabular}
\caption{\label{tab:surveyresults} Results of the student survey, coded according to the categories in Table~\ref{tab:surveyquestions}. }
\vspace{-0.00in}
\end{table}

In this work we study the problem of student evaluation at scale via peer grading using ordinal feedback.
We cast this as a rank aggregation problem and study different probabilistic models for obtaining student grades, as well as estimating the reliability of the peer graders.
Using data collected from a real course, we find that the performance of ordinal peer grading methods is at least competitive with cardinal methods for grade estimation, even though they require strictly less information from the graders. For grader reliability estimation, Mallow's model outperforms all other methods, and it shows consistently good and robust performance for grade estimation as well.
In general, we find that ordinal peer grading is robust and scalable, offering a grading accuracy that is comparable to TA grading in our course.


\bibliographystyle{abbrv}
\begin{small}
\bibliography{peer-assesment}

\begin{thebibliography}{10}

\bibitem{ailon_mallowsmle_2008}
N.~Ailon, M.~Charikar, and A.~Newman.
\newblock Aggregating inconsistent information: Ranking and clustering.
\newblock {\em J. ACM}, 55(5):23:1--23:27, Nov. 2008.

\bibitem{arrow_socialchoice_1970}
K.~J. Arrow.
\newblock {\em {Social Choice and Individual Values}}.
\newblock Yale University Press, 2nd edition, Sept. 1970.

\bibitem{aslam_metasearch_2001}
J.~A. Aslam and M.~Montague.
\newblock Models for metasearch.
\newblock In {\em SIGIR}, pages 276--284, 2001.

\bibitem{bachrach_testgrade_2012}
Y.~Bachrach, T.~Graepel, T.~Minka, and J.~Guiver.
\newblock How to grade a test without knowing the answers - a bayesian
  graphical model for adaptive crowdsourcing and aptitude testing.
\newblock In {\em ICML}, 2012.

\bibitem{barnett_2003}
W.~Barnett.
\newblock The modern theory of consumer behavior: Ordinal or cardinal?
\newblock {\em The Quarterly Journal of Austrian Economics}, 6(1):41--65, 2003.

\bibitem{bashir_rankaggreg_2013}
M.~Bashir, J.~Anderton, J.~Wu, P.~B. Golbus, V.~Pavlu, and J.~A. Aslam.
\newblock A document rating system for preference judgements.
\newblock In {\em SIGIR}, pages 909--912, 2013.

\bibitem{bouzidi_2009}
L.~Bouzidi and A.~Jaillet.
\newblock Can online peer assessment be trusted?
\newblock {\em Educational Technology \& Society}, 12(4):257--268, 2009.

\bibitem{bradley_1952}
R.~A. Bradley and M.~E. Terry.
\newblock Rank analysis of incomplete block designs: I. the method of paired
  comparisons.
\newblock {\em Biometrika}, 39(3/4):pp. 324--345, 1952.

\bibitem{carterette_2008}
B.~Carterette, P.~N. Bennett, D.~M. Chickering, and S.~T. Dumais.
\newblock Here or there: Preference judgments for relevance.
\newblock In {\em ECIR}, pages 16--27, 2008.

\bibitem{chang_2011}
C.-C. Chang, K.-H. Tseng, P.-N. Chou, and Y.-H. Chen.
\newblock Reliability and validity of web-based portfolio peer assessment: A
  case study for a senior high school's students taking computer course.
\newblock {\em Comput. Educ.}, 57(1):1306--1316, Aug. 2011.

\bibitem{chen_rankaggreg_2013}
X.~Chen, P.~N. Bennett, K.~Collins-Thompson, and E.~Horvitz.
\newblock Pairwise ranking aggregation in a crowdsourced setting.
\newblock In {\em WSDM}, pages 193--202, 2013.

\bibitem{diaz_nipsworkshop_2013}
J.~Diez, O.~Luaces, A.~Alonso-Betanzos, A.~Troncoso, and A.~Bahamonde.
\newblock Peer assessment in moocs using preference learning via matrix
  factorization, 2013.

\bibitem{dwork_rankagg_2001}
C.~Dwork, R.~Kumar, M.~Naor, and D.~Sivakumar.
\newblock Rank aggregation methods for the web.
\newblock In {\em WWW}, pages 613--622, 2001.

\bibitem{freeman_2010}
S.~Freeman and J.~W. Parks.
\newblock How accurate is peer grading?
\newblock {\em CBE-Life Sciences Education}, 9(4):482--488, 2010.

\bibitem{guiver_placlucebayes_2009}
J.~Guiver and E.~Snelson.
\newblock Bayesian inference for plackett-luce ranking models.
\newblock In {\em ICML}, pages 377--384, 2009.

\bibitem{jonathanhaber_assessment}
J.~Haber.
\newblock
  \url{http://degreeoffreedom.org/between-two-worlds-moocs-and-assessment}.

\bibitem{jonathanhaber_screwing}
J.~Haber.
\newblock \url{http://degreeoffreedom.org/mooc-assignments-screwing/}, Oct.
  2013.

\bibitem{herbrich_trueskill_2007}
R.~Herbrich, T.~Minka, and T.~Graepel.
\newblock Trueskill$^{\mbox{tm}}$: A bayesian skill rating system.
\newblock In {\em NIPS}, pages 569--576, 2007.

\bibitem{ipeirotis_crowdsourcing_2011}
P.~G. Ipeirotis and P.~K. Paritosh.
\newblock Managing crowdsourced human computation: a tutorial.
\newblock In {\em WWW}, pages 287--288, 2011.

\bibitem{kendall_rankcorrelation_1948}
M.~Kendall.
\newblock {\em Rank correlation methods}.
\newblock Griffin, London, 1948.

\bibitem{kenyon-mathieu_mallowsmle_2007}
C.~Kenyon-Mathieu and W.~Schudy.
\newblock How to rank with few errors.
\newblock In {\em STOC}, pages 95--103, 2007.

\bibitem{kulkarni:2013}
C.~Kulkarni, K.~Wei, H.~Le, D.~Chia, K.~Papadopoulos, J.~Cheng, D.~Koller, and
  S.~Klemmer.
\newblock Peer and self assessment in massive online classes.
\newblock {\em ACM Trans. CHI}, 20(6):33:1--33:31, Dec. 2013.

\bibitem{lebanon_cranking_2002}
G.~Lebanon and J.~D. Lafferty.
\newblock Cranking: Combining rankings using conditional probability models on
  permutations.
\newblock In {\em ICML}, pages 363--370, 2002.

\bibitem{liu_learningtorank_2009}
T.-Y. Liu.
\newblock Learning to rank for information retrieval.
\newblock {\em Found. Trends Inf. Retr.}, 3(3):225--331, Mar. 2009.

\bibitem{lu_mallows_2011}
T.~Lu and C.~Boutilier.
\newblock Learning mallows models with pairwise preferences.
\newblock In {\em ICML}, pages 145--152, June 2011.

\bibitem{lu_socialchoice_2010}
T.~Lu and C.~E. Boutilier.
\newblock The unavailable candidate model: A decision-theoretic view of social
  choice.
\newblock In {\em EC}, pages 263--274, 2010.

\bibitem{luce_1961}
R.~D. Luce.
\newblock {\em {Individual Choice Behavior: A theoretical analysis}}.
\newblock Wiley, 1959.

\bibitem{mallows_1957}
C.~L. Mallows.
\newblock Non-null ranking models.
\newblock {\em Biometrika}, 44(1/2):pp. 114--130, 1957.

\bibitem{miller_1956}
G.~A. Miller.
\newblock The magical number seven, plus or minus two: Some limits on our
  capacity for processing information.
\newblock {\em The Psychological Review}, 63(2):81--97, March 1956.

\bibitem{mosfert_2013}
M.~Mostert and J.~D. Snowball.
\newblock Where angels fear to tread: online peer-assessment in a large
  first-year class.
\newblock {\em Assessment \& Evaluation in Higher Education}, 38(6):674--686,
  2013.

\bibitem{niu_stochrankagg_2013}
S.~Niu, Y.~Lan, J.~Guo, and X.~Cheng.
\newblock Stochastic rank aggregation.
\newblock {\em CoRR}, abs/1309.6852, 2013.

\bibitem{piech_peergrading_2013}
C.~Piech, J.~Huang, Z.~Chen, C.~Do, A.~Ng, and D.~Koller.
\newblock Tuned models of peer assessment in {MOOC}s.
\newblock In {\em EDM}, 2013.

\bibitem{plackett_1975}
R.~L. Plackett.
\newblock The analysis of permutations.
\newblock {\em Journal of the Royal Statistical Society. Series C (Applied
  Statistics)}, 24(2):193--202, 1975.

\bibitem{qin_cpsrankagg_2010}
T.~Qin, X.~Geng, and T.-Y. Liu.
\newblock A new probabilistic model for rank aggregation.
\newblock In {\em NIPS}, pages 1948--1956, 2010.

\bibitem{raykar_learningfromcrowds_2010}
V.~C. Raykar, S.~Yu, L.~H. Zhao, G.~H. Valadez, C.~Florin, L.~Bogoni, and
  L.~Moy.
\newblock Learning from crowds.
\newblock {\em JMLR}, 11:1297--1322, Aug. 2010.

\bibitem{jonathanrees_peergradingcriticism}
J.~Rees.
\newblock
  \url{http://www.insidehighered.com/views/2013/03/05/essays-flaws-peer-grading-moocs}.

\bibitem{shah_2013}
N.~Shah, J.~Bradley, A.~Parekh, M.~Wainwright, and K.~Ramchandran.
\newblock A case for ordinal peer-evaluation in {MOOC}s, 2013.

\bibitem{stewart_2005}
N.~Stewart, G.~D.~A. Brown, and N.~Chater.
\newblock Absolute identification by relative judgment.
\newblock {\em Psychological Review}, 112:881--911, 2005.

\bibitem{thurstone_1927}
L.~L. Thurstone.
\newblock The method of paired comparisons for social values.
\newblock {\em Journal of Abnormal and Social Psychology}, 27:384--400, 1927.

\bibitem{volkovs_prefaggr_2012}
M.~N. Volkovs and R.~S. Zemel.
\newblock A flexible generative model for preference aggregation.
\newblock In {\em WWW}, pages 479--488, 2012.

\end{thebibliography}
\end{small}

\end{document}